\documentclass[twocolumn,twoside]{svjour3}
\smartqed
\usepackage{booktabs}
\usepackage{times}
\usepackage{epsfig}
\usepackage{epstopdf}
\usepackage{amsmath}
\usepackage{amssymb}
\usepackage{caption}
\usepackage{tabularx}
\usepackage[table]{xcolor}
\usepackage[normalem]{ulem}
\usepackage[noend]{algpseudocode}
\usepackage{lipsum}
\usepackage[cmintegrals]{newtxmath}
\usepackage[colorlinks]{hyperref}

\makeatletter
\DeclareRobustCommand\onedot{\futurelet\@let@token\@onedot}
\def\@onedot{\ifx\@let@token.\else.\null\fi\xspace}

\makeatother

\newcolumntype{L}[1]{>{\raggedright\arraybackslash}p{#1}}
\newcolumntype{C}[1]{>{\centering\arraybackslash}p{#1}}
\newcolumntype{R}[1]{>{\raggedleft\arraybackslash}p{#1}}

\usepackage{url}
\usepackage[all]{genealogytree}
\usepackage[framemethod=tikz]{mdframed}
\usepackage{hyperref}
\hypersetup{colorlinks = true, 
	    linkcolor = black, 
	    urlcolor = blue,
            citecolor = blue} 

\usepackage{xcolor}
\usepackage{mdframed}
\mdfdefinestyle{mystyle}{
    backgroundcolor=yellow!15,
    linewidth=0.5pt
}
\usepackage{xspace}
\usepackage{subfig}
\usepackage{array}
\usepackage{soul}

\usepackage{natbib}
\usepackage{graphicx}%
\usepackage{multirow}%
\usepackage{amsmath,amssymb,amsfonts}%
\usepackage{mathrsfs}%
\usepackage[title]{appendix}%
\usepackage{xcolor}%
\usepackage{textcomp}%
\usepackage{manyfoot}%
\usepackage{booktabs}%
\usepackage{algorithm}%
\usepackage{algorithmicx}%
\usepackage{algpseudocode}%
\usepackage{listings}%
\newcommand{\revise}[1]{\textcolor{black}{#1}}

\usepackage{url}

\usepackage[all]{genealogytree}
\usepackage[framemethod=tikz]{mdframed}

\usepackage{hyperref}
\hypersetup{colorlinks = true, 
	    linkcolor = black, 
	    urlcolor = blue,
            citecolor = blue}

\usepackage{xcolor}
\usepackage{mdframed}
\mdfdefinestyle{mystyle}{
    backgroundcolor=yellow!15,
    linewidth=0.5pt
}

\usepackage{booktabs}      
\usepackage{graphicx}
\usepackage{amsmath}
        
\usepackage{amssymb}
\usepackage{xspace}
\usepackage{multirow}
\usepackage{subfig}
\usepackage{array}
\usepackage{soul}

\begin{document}

\title{On the Trustworthiness Landscape of State-of-the-art Generative Models: A Survey and Outlook}

\author{$\text{Mingyuan~Fan}^{1\star}$ \and
        $\text{Chengyu~Wang}^{2}$ \and
        $\text{Cen~Chen}^{1\star}$ \and
        $\text{Yang~Liu}^{3}$ \and
        $\text{Jun~Huang}^{2}$ \\
        \email{\{fmy2660966,~~chywang2013,~~cecilia.cenchen\}@gmail.com, \\
        bcds2018@foxmail.com, huangjun.hj@alibaba-inc.com
        }}

\authorrunning{Mingyuan Fan et al.} 

\institute{$~^1$ East China Normal University, Shanghai, China. \\
$~^2$ Alibaba Group, Hangzhou, China.\\
$~^3$ Xidian University, Xian, China. \\
$\star$ Corresponding authors.}

\date{Received: date / Accepted: date}

\maketitle

\begin{abstract}
Diffusion models and large language models have emerged as leading-edge generative models, revolutionizing various aspects of human life. However, their practical implementation has also exposed inherent risks, bringing to light their potential downsides and sparking concerns about their trustworthiness. Despite the wealth of literature on this subject, a comprehensive survey that specifically delves into the intersection of large-scale generative models and their trustworthiness remains largely absent. To bridge this gap, this paper investigates both long-standing and emerging threats associated with these models across four fundamental dimensions: 1) privacy, 2) security, 3) fairness, and 4) responsibility. Based on our investigation results, we develop an extensive survey that outlines the trustworthiness of large generative models. Following that, we provide practical recommendations and identify promising research directions for generative AI, ultimately promoting the trustworthiness of these models and benefiting society as a whole.

\end{abstract}

\maketitle

\newtheorem{motivating_example}{Motivating Example}
\newcommand{\citez}[1]{\citet{#1}}

\section{Introduction}
\label{section:intro}

The utilization of diffusion models (DMs)~\citep{diffusion2,DALLE} and large language models (LLMs)~\citep{gpt4} has surged across various real-world applications, enabling the generation of content that rivals human expertise.
For instance GPT-4 has become a ubiquitous productivity tool worldwide, offering invaluable assistance across diverse domains—from serving as a virtual assistant to generating code segments for engineers~\citep{gpt3,gpt4}.
Moreover, multimodal models built upon DMs and LLMs have achieved significant advancements in transforming content from one modality to another, particularly in bridging language and vision~\citep{DALLE2,image_editing,video_generation}.

\begin{figure}
    \centering
    \includegraphics[width=0.99\linewidth]{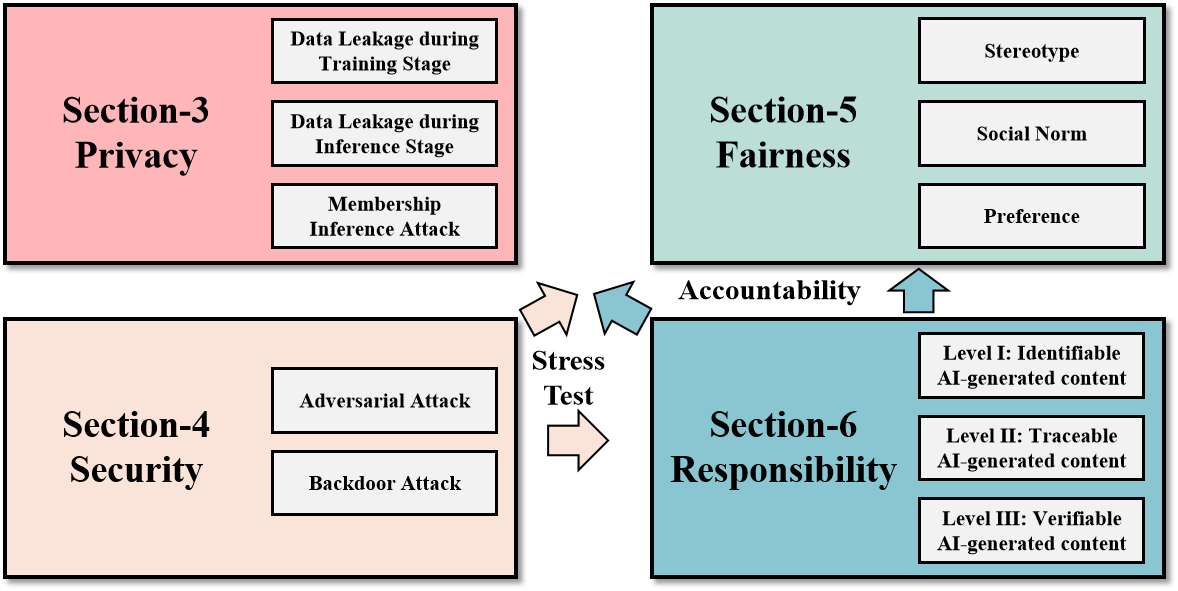}
    \caption{The trustworthiness landscape of DMs and LLMs.}
    \label{paper_overview}
\end{figure}

While these models offer substantial social benefits, their malicious exploitation~\citep{data_leakage_4,data_leakage_2} has raised significant concerns about their trustworthiness.
DMs have been criticized for exacerbating societal divisions, as the images they generate may revive harmful stereotypes or manipulate public opinion.
For example, in April 2023, an organization misused DMs to generate \hyperlink{https://www.bloomberg.com/graphics/2023-generative-ai-bias/}{misleading information} for specific agendas.
Similarly, LLMs have been implicated in serious issues, including contributing to \hyperlink{https://www.vice.com/en/article/man-dies-by-suicide-after-talking-with-ai-chatbot-widow-says/}{suicide cases}, \hyperlink{https://www.bbc.com/news/world-us-canada-65735769}{fabricating legal cases}, and \hyperlink{https://www.engadget.com/chatgpt-briefly-went-offline-after-a-bug-revealed-user-chat-histories-115632504.html}{leaking users' chat histories}.
This troubling trend is expected to escalate, with \hyperlink{https://www.europol.europa.eu/cms/sites/default/files/documents/Europol_Innovation_Lab_Facing_Reality_Law_Enforcement_And_The_Challenge_Of_Deepfakes.pdf}{an official academic institution} suggesting that within a few years, 90\% of online content could be generated by these models.
\hyperlink{https://www.weforum.org/publications/global-risks-report-2024/in-full/}{A report} from the World Economic Forum predicts that such content will completely reshape public perception in the near future.
In response, the trustworthiness landscape of DMs and LLMs is evolving rapidly, with numerous initiatives underway.
However, a significant gap remains in systematically organizing and critically reviewing these efforts.
To fill this gap, as shown in Figure~\ref{paper_overview}, we embark on a systematic trustworthiness investigation by organizing recent advances around four fundamental dimensions: privacy, security, fairness, and responsibility:

\begin{itemize}
    \item \textbf{Privacy (Section~\ref{section:privacy}).} 
    Developing privacy-preserving models has gained global consensus~\citep{data_leakage_2}.
    The implications of privacy leakage are far-reaching, leading to diminished user trust, malicious outcomes, and potential regulatory violations.
	DMs and LLMs are particularly vulnerable to sensitive data leakage~\citep{data_leakage_2,data_leakage_4}, as they can directly capture the underlying distribution of training data.
    We investigate the issue of privacy leakage in DMs and LLMs throughout the training and inference phases, as well as membership inference attacks which determine whether given data points were part of the training set.

    \item \textbf{Security (Section~\ref{section:security}).}
    Ensuring the robustness of DMs and LLMs against malicious attacks is critical for their real-world deployment.
    Two typical forms of attack are adversarial attacks~\citep{adv_1} and backdoor attacks~\citep{backdoor_1,wiper}.
    Adversarial attacks exploit a model's inherent vulnerabilities through minor input modifications.
    Backdoor attacks insert a hidden backdoor into the model, which, when activated during inference, causes the model to behave unpredictably.
    Both attack types can manipulate the model or significantly deteriorate its performance.
    
    \item \textbf{Fairness (Section~\ref{section:fairness}).}
    As DMs and LLMs increasingly influence our daily lives, maintaining the principle of fairness to ensure equitable treatment across all social segments is crucial.
    These models should operate within ethical and moral frameworks to avoid the perpetuation of prejudice and societal division.
    However, AI-generated content often exhibits biases~\citep{fairness_3,new_fairness_14}, resulting in unfair outcomes and discrimination against specific social groups.
    We review recent advancements in improving fairness in DMs and LLMs through three lenses: stereotype, social norm, and preference.

    \item \textbf{Responsibility (Section~\ref{section:responsibility}).}
    The responsibility of DMs and LLMs encompasses the duty to proactively prevent misuse and mitigate potential disruptions to societal norms.
    We categorize responsibility into three progressively refined tiers for review: identifiability, traceability, and verifiability, each presenting increasing levels of implementation complexity.
    Identifiability pertains to the ability to distinguish between human-created and AI-generated content.
    Achieving this can significantly reduce the likelihood of social rumors and similar incidents.
    Traceability requires models to explicitly embed watermarks in their generated content, facilitating accountability by tracing content back to the respective AI model.
    Verifiability involves the authentication of AI-generated content, thereby enhancing users' trust in model decisions.
\end{itemize}
The four dimensions are intricately connected and interdependent, each addressing unique facets while reinforcing one another.
Security evaluates a model's resilience under extreme conditions by leveraging adversarial and backdoor attacks.
These attacks can serve as stress tests in the contexts of fairness and responsibility, exposing whether the model operates impartially, ethically, and accountably.
Responsibility underpins both fairness and privacy by ensuring oversight and accountability, so as to drive the commitment to ethical practices in model deployment.

\textbf{Distinct features of this survey.}
\citet{diffusion_model_survey} reviewed the trustworthiness of DMs, while \citet{related_survey_2} evaluated LLM alignment with human behavior.
\citet{related_survey_3} scrutinized existing research from a benchmarking perspective. 
In contrast, our survey extends beyond the scope of these papers in two significant ways.
First, it expands the horizon of current surveys by amalgamating insights on both DMs and LLMs, aligning with the prevailing trend of integrating these as multimodal models.
By comparing the trustworthiness of LLMs and DMs, we aim to deepen the understanding of the distinct characteristics inherent in each modality, fostering interdisciplinary dialogue and encouraging the exchange of methodologies and theoretical insights.
Second, this survey goes beyond engineering-focused metrics and standardized benchmarks to provide a comprehensive understanding of the development and evolution of trustworthiness in DMs and LLMs, advocating for flexible evaluation procedures that account for regional and temporal variations.
In summary, this survey yields four key benefits:
\begin{itemize}
    \item \textbf{A Panoramic Overview:} This survey provides a comprehensive view of trustworthiness in the context of DMs and LLMs, offering a holistic perspective.
    \item \textbf{New Taxonomy:} A novel classification framework is introduced, aimed at structuring the existing body of research on fairness and responsibility\footnote{Security and privacy have been extensively studied; for these aspects, we adopt a widely recognized framework.}. Our taxonomy groups fairness into three areas: stereotypes, social norms, and preferences, while responsibility is structured into three tiers: identifiability, traceability, and verifiability.
    \item \textbf{Industry Risk Awareness:} The survey highlights potential risks associated with deploying these models in real-world settings and offers valuable insights for industry practitioners, including potential strategies.
    \item \textbf{Future Directions:} The survey identifies promising areas and untapped opportunities that are ripe for further exploration, aiming to catalyze future research efforts.
\end{itemize}

\textbf{Roadmap.}
Section~\ref{section:models} offers an overview of DMs and LLMs, laying the groundwork for the subsequent sections.
Sections~\ref{section:privacy}, \ref{section:security}, \ref{section:fairness}, and \ref{section:responsibility} delve into the four dimensions in detail, beginning with motivating examples and followed by a review of advancements in each area.
This is complemented by a benchmark tool subsection and a discussion subsection, which review existing evaluation metrics and datasets and summarize key insights and opportunities for further research, respectively.
Section~\ref{section:conclusion} wraps up the paper.

\section{A Glimpse of State-of-the-art Generative Models}
\label{section:models}

\textbf{DMs.}
At the heart of DMs lies the diffusion process, inspired by non-equilibrium thermodynamics \citep{diffusion,score_dm}, which gradually converts a simple distribution, typically Gaussian noise, into a complex one.
Formally, given a natural sample $x_0$, the transition between two consecutive diffusion steps is defined as:
\begin{equation}
\label{diffusion_process}
q(x_t|x_{t-1}) = \mathcal{N}(x_t; \sqrt{\alpha_t} x_{t-1}, (1-\alpha_t) \textbf{I} ), t=1,\cdots,T,
\end{equation}
where $\alpha_t \in (0, 1)$ is a noise schedule parameter to control the amount of noise added at each step.
Equation \ref{diffusion_process} can be simplified using the reparameterization trick $x_t = \sqrt{\overline{\alpha}_t} x_0 + \sqrt{1-\alpha_t} \epsilon_0$ where $\sqrt{\overline{\alpha}_t} = \prod_{i=1}^t \alpha_i$ and $\epsilon_0 \sim \mathcal{N}(0, \textbf{I})$.
DMs, denoted as $\epsilon_{\theta}$, are trained to predict the original image $x_0$ from its noisy version by minimizing the difference between the actual noise added and the noise predicted by the model, i.e., $\mathbb{E}_{x_0, \epsilon_0} || \epsilon_0 - \epsilon_{\theta}(x_t, t)  ||_2^2.$
Intuitively, DMs predict what noise can enhance the natural appearance of a given noisy image.
Once trained, DMs can perform a $T$-step denoising process on a Gaussian noise $\epsilon$ to generate realistic images:
\begin{equation}
\label{sampling_process}
x_{t-1} = \frac{1}{\sqrt{\alpha_t}} \left(x_t - \frac{1-\alpha_t}{\sqrt{1-\overline{\alpha}_t}}  \epsilon_{\theta}(x_t, t) \right) + \frac{(1-\alpha_t)(1-\overline{\alpha}_{t-1})}{1-\overline{\alpha}_{t}} \epsilon.
\end{equation}
\revise{
This iterative adjustment simplifies the learning process compared to generating an image in a single step, allowing for richer textures and more intricate details in the final outputs~\citep{Imagen}.
The above formulation is known as denoising diffusion probabilistic models (DDPM).
However, the denoising process is time-consuming due to the large number of steps required.
To improve efficiency, a variation called denoising diffusion implicit models (DDIM)~\citep{ddim} adopts a non-Markovian approach, enabling the model to skip certain steps.
Other recent advances include consistency model~\citep{consistency_model} and rectified flow~\citep{rectified_flow}.
The former learns to predict the clean data from noisy samples directly at each diffusion step, while the latter regularizes the denoising trajectory to make it simpler and more sampling-efficient.
Some research proposed performing the diffusion and denoising processes in latent space instead of pixel space~\citep{stable_diffusion}.
}
Moreover, while vanilla DMs are designed for unconditional image generation~\citep{diffusion}, they can be extended to conditional tasks \citep{DALLE} with additional supervision signals $y$ to enhance their adaptability and versatility~\citep{video_generation,image_coloring}.
This requires an encoder $\tau_w$ to map $y$ to a latent vector, which is incorporated into the diffusion and denoising processes, i.e., modifying $\epsilon_{\theta}(x_t, t)$ to $\epsilon_{\theta}(x_t, \tau_w(y), t)$.
Among these supervision signals, labels and textual descriptions~\citep{DALLE} are particularly significant, enabling DMs to generate images that align with the provided descriptions.
For tasks like image coloring, supervision may also come in the form of images, where the model transforms a gray-scale source image into its colored counterpart~\citep{image_coloring}.

\textbf{LLMs.}
LLMs~\citep{gpt1,gpt2} predict the next word for a given prefix or fill in masked portions of text within a specific context.
Their foundation lies in the Transformer architecture~\citep{transformer}, a significant milestone in natural language processing.
Building upon this architecture, subsequent LLMs~\citep{gpt3,gpt4,LAMDA,OPT} have incorporated recent advancements, such as prompt learning~\citep{prompt_learning}, to further enhance their capabilities.
Prompt learning~\citep{prompt_learning_1,prompt_learning_2} involves inserting a prompt before the original input, activating specific latent patterns within LLMs and enabling them to concentrate on task-specific skills.
Other related concepts, such as adapters~\citep{adapter}, instruction learning~\citep{instruction_learning}, in-context learning~\citep{in_context_learning}, and human alignment approaches~\citep{gpt4} also contribute significantly to the effectiveness of LLMs.
Intuitively, LLMs can learn general knowledge from vast text corpora; prompts, adapters, and similar ones can be viewed as external parameters to store task-specific knowledge.
These techniques are parameter-efficient, reducing both training costs and the risk of catastrophic forgetting and overfitting.
A recent development, retrieval-augmented generation (RAG) \citep{model_part_rag_1}, introduces a retriever to extract relevant information from a pre-built knowledge base according to user queries, aiding LLMs in generating responses.
RAG allows knowledge to be stored externally rather than within model parameters, enhancing inference efficiency and reducing the risk of sensitive information leakage, while also enabling updates to the knowledge base for reliable, up-to-date information.
Advanced RAG techniques \citep{model_part_rag_2} involve multi-granularity retrieval, retrieval refinements, as well as multi-round and sequential retrievals for complex queries.
These advancements have propelled LLMs to performance levels comparable to human experts in numerous tasks~\citep{gpt4}.

\textbf{DMs vs. LLMs.}
DMs and LLMs share some similarities, yet they encounter distinct challenges regarding trustworthiness due to differences in data modalities (images vs. text) and their learning and inference processes (diffusion vs. token prediction).
Firstly, images are continuous data and are less affected by minor perturbations that do not change their semantic meaning.
In contrast, small alterations in text, such as substituting a single word, can fundamentally alter the entire sentence's meaning.
Even when text is represented in continuous space through embeddings, converting these embeddings back into words can lead to grammatical and approximation errors.
Furthermore, text, being a condensed form of human expression, inherently conveys emotions more directly than images, which tend to contain more redundant information.
Secondly, both DMs and LLMs process data iteratively, but they differ in how they approach each iteration.
DMs update all pixels of an image simultaneously, with each pixel‘s update depending on the others.
Conversely, LLMs predict only one token at a time, relying on the surrounding context while leaving the rest of the input unchanged.
This means that LLMs may face sharp breaks in logic or flow when a token prediction goes wrong.
Additionally, LLMs have variable output dimensions, whereas DMs produce outputs of fixed dimensions.
Many trustworthiness-related issues can be framed as optimization tasks, such as data reconstruction.
These differences necessitate tailored optimization strategies for each model.

\section{Privacy}
\label{section:privacy}

\begin{mdframed}[style=mystyle]
\begin{motivating_example}
\label{privacy_case}
\textbf{(Case for Privacy Leakage in Real-life and Its Impact)}
\textit{
Stable Diffusion, an open-source AI art generator developed by Stability AI, has become embroiled in a legal dispute. In 2023, Getty Images initiated legal proceedings against Stability AI, Inc. by filing a complaint in the United States District Court in Delaware. This action was taken after Getty Images alleged that Stable Diffusion inadvertently leaked its training data during the inference stage, which included watermarked images from Getty Images' collection. It is suspected that the number of images resembling those in Getty Images' database may exceed a staggering 12 million.
[\hyperlink{https://www.theverge.com/2023/2/6/23587393/ai-art-copyright-lawsuit-getty-images-stable-diffusion}{Link}]
}
\end{motivating_example}
\end{mdframed}
The extensive number of parameters in DMs and LLMs enables them to learn from vast training data.
However, this over-parameterization can unintentionally create shortcuts that allow models to achieve high performance by merely memorizing training samples, leading to privacy leakage.
Example~\ref{privacy_case} illustrates a real-world incident of privacy leakage involving a DM, inciting significant criticism from Getty Images, the owner of some of the implicated training data.
This underscores the urgent need to address privacy issues when deploying models.
We review recent developments concerning the privacy issues of DMs and LLMs.

\subsection{Overview}
A model is considered privacy-preserving if it safeguards information about its training data throughout its entire lifecycle, in a way that no feasible methods exist to derive such information\footnote{\url{https://www.dlapiperdataprotection.com/}}.
Owing to the distinct differences between training and inference stages, we examine data leakage concerns separately for each phase.
During the training stage, we focus on federated learning and split learning, while in Section~\ref{MIA}, we specifically address membership inference attacks targeting DMs and LLMs, which seek to determine the membership status of particular data.

\subsection{Data Leakage during Training Stage}
\label{data_leakage_training}
Two paradigms exist for training deep neural networks: centralized training and distributed training with multi-party participation~\citep{fl_survey}.
Centralized training provides better data protection, particularly with strong access control.
Conversely, the latter can heighten the risk of privacy leakage due to the involvement of untrusted parties.
Federated learning and split learning are two key collaborative frameworks.
Unless specified, we do not distinguish between DMs and LLMs in this context, as both frameworks can be applied to either model type.

\subsubsection{Federated Learning}
In federated learning, a server distributes a globally shared model to clients, who perform local computations on their data to generate gradients.
These gradients are sent back to be aggregated to update the global model.
While federated learning avoids sharing raw data, gradient leakage attacks can still reconstruct clients' data from shared gradients.

\textbf{Gradient Leakage Attacks.}
Gradient leakage attacks~\citep{gla,igla} exploit gradient matching techniques to recover clients' data by aligning uploaded gradients with those from dummy data, enabling high-fidelity reconstruction.
Subsequent research~\citep{invert_grad,grad_inversion,evaluate_gla} has improved the efficacy of these attacks by utilizing cosine similarity loss functions, adding regularization terms, and refining initialization methods, among others.
Interestingly, optimized dummy images often resemble random noise, which hinders attack performance~\citep{grad_image_prior}.
To address this, \citet{grad_image_prior} and \citet{false_sense} optimized generative models' latent vectors to generate clearer images.
\citet{tag} found that combining $L_1$ and $L_2$ loss functions can improve text reconstruction.
\citet{bayes_attack_for_gla} proposed a unified Bayesian framework for gradient leakage attacks, noting that larger models are more susceptible to privacy leakage.
Besides, DMs and LLMs tend to be more susceptible to gradient leakage attacks because they directly model the training data.
For example, \citet{recovering_private_text_llm} leveraged the generative capabilities of these models to generate candidate data and determined the optimal reconstruction by comparing the similarity between candidate gradients and the uploaded gradients.
As model performance improves, so does the quality of reconstruction.

\textbf{Gradient Leakage Defenses.}
Encryption and perturbation are the primary techniques used to defend against gradient leakage attacks.
The encryption methods~\citep{crypt_1,crypt_2} prevent attackers from accessing individual client gradients, allowing only access to the plaintext of aggregated gradients, which forces them to reconstruct all clients' data at once and complicates gradient matching problem.
However, encryption methods can introduce a high computational overhead, limiting their usability in resource-constrained settings~\citep{soteria,false_sense}.
Perturbation-based methods apply slight gradient modifications to confound attackers.
Differential privacy~\citep{dp} injects random noise into gradients, while Top-K gradient sparsification~\citep{gla} retains only the most significant gradient elements.
Gradient quantization~\citep{false_sense} represents the gradients with lower-bit precision, reducing the amount of sensitive information.
Selective pruning evaluates the private information within each gradient element and then implements gradient pruning~\citep{soteria}.
Another approach is to generate data that contains less private information, typically by optimizing metrics related to data privacy and model utility~\citep{refiner}.

\textbf{Remark.}
Gradient leakage attacks, initially designed for discriminative models, can also be applied to generative models, with DMs and LLMs being particularly vulnerable due to their ability to replicate training data patterns.
As models become more proficient, the risk of disclosing training data also increases.
In light of this, one possible mitigation involves commencing training with sensitive data and switching to less sensitive data; however, the impact of this strategy on model performance remains uncertain.
For LLMs, RAG can be employed to store privacy-sensitive information locally, keeping it out of the training set for privacy protection.
Furthermore, fine-tuning LLMs with adapters, rather than training from scratch, allows for a restricted sharing of gradients, reducing the risk of privacy breaches.
Nonetheless, the effectiveness of the above several mitigations in counterbalancing the privacy risks introduced by models' generation capabilities remains unclear.

\subsubsection{Split Learning}
In split learning, a model is divided into two sub-models: the bottom network on the client side and the top network on the server side.
Clients process input data through the bottom network, sending intermediate results to the server for further computation using the top network.
Privacy concerns arise when one party is untrustworthy.

\textbf{Data Leakage.}
\citet{split_attack_1} introduced Feature Space Hijacking Attack (FSHA), targeting the reconstruction of client data when the label party is untrustworthy.
In FSHA, the adversarial label party trains a shadow network to mimic the outputs of client's network, and then a decoder maps these outputs back to their corresponding data.
By applying outputs from client's bottom network to the trained decoder, client data can be revealed.
\citet{new_split_learning_data_leakage_1} relaxed FSHA's requirements, allowing the use of common public data instead of data similar to client data.
\citet{new_split_learning_data_leakage_2} showed that replacing the decoder with a diffusion model improves reconstruction quality.

Existing defense measures, such as detection and perturbation, are found to be ineffective against FSHA~\citep{split_attack_1}.
To address this, \citet{ResSFL} developed ResSFL, which trains a feature extractor resistant to inversion and initializes the bottom model with it.
\citet{new_split_learning_data_leakage_3} theoretically analyzed the decoder’s reconstruction limits using Fisher information.
\citet{new_split_learning_data_leakage_4} employed a regularization term to reduce the correlation between the input data and intermediate activation values, alongside pruning sensitive parameters for defense.

\textbf{Label Leakage.}
\citet{label_leakage_1} showed that backward gradient sign patterns from the top model can leak labels and suggested using gradient perturbations to mitigate.
However, \citet{split_learning_2} argued against the effectiveness of this method and advocated for multiple activations and label mixing instead.
\citet{split_attack_2} and \citet{split_attack_3} showed that labels could be inferred with gradient matching, and \citet{BlindFL} found that even a small amount of labeled samples is sufficient to fine-tune models, enabling direct label predictions for training data.
\citet{PSLF} proposed using flipped labels to compute gradients to obfuscate attackers while training a private sub-model with true labels to compensate for performance losses.

\textbf{Remark.}
Split learning restricts direct access to the complete model by any single entity and can intuitively impede attackers from leveraging the generative capacity of DMs and LLMs to reconstruct data.
In this context, the generative potential of DMs and LLMs does not necessarily exacerbate privacy leakage.
Nevertheless, the certainty of this claim is still in question, as the possibility of attackers developing shadow models to compensate for this restriction persists, highlighting the necessity for further research.

\subsection{Data Leakage during Inference Stage}
\label{data_leakage_inference}

During the inference stage, attackers can craft specific inputs to prompt models into producing outputs that reveal aspects of the training data, resulting in privacy leakage.
Notably, multi-modal DMs tend to leak images rather than the text prompts used to generate those images.
Furthermore, the types of data leaked by DMs and LLMs exhibit significant differences.
LLMs are particularly prone to leaking entity relationships, such as names, locations, and email addresses, but they fail to remember numerical information like phone numbers.
In contrast, DMs appear to memorize and reproduce training images without favoring specific types.
Moreover, the implications of data leakage differ for these models.
DMs encounter issues like copyright and portrait rights, whereas LLMs face the potential exposure of personal contact information and intricate entity relationships.

\subsubsection{DMs}

\textbf{Attacks.}
\citet{data_leakage_1,data_leakage_2,data_leakage_3} first investigated the issue of training data leakage in DMs, including Stable Diffusion and Imagen.
The empirical studies in~\citep{data_leakage_1,data_leakage_3} revealed that DMs could memorize and reproduce various elements of the training data as outputs during inference.
Complementing this, \citet{data_leakage_2} utilized a brute-force method to verify the occurrence of data replication in DMs.
By inputting a wide array of prompts with different random seeds, they generated a large volume of data and then applied a filtering process, revealing numerous instances of replicated images that bore a striking resemblance to the training data, some at a near pixel-perfect level.

\textbf{Defense.}
To mitigate the memorization of training data, a simple solution is to deduplicate the training dataset, a measure validated by OpenAI as effective for \hyperlink{https://openai.com/index/dall-e-2-pre-training-mitigations/}{DALLE2}.
Interestingly, \citet{privacy_leakage_inference_dm} demonstrated that even with repeated images, the memorization effect can be substantially reduced if image captions remain sufficiently diverse.
They then suggested rewriting captions during the training phase or introducing noise to user inputs during the inference phase.
Building on these findings, \citet{privacy_leakage_inference_dm2} introduced anti-memorization techniques, guiding DMs away from training data during image generation through despecification, caption deduplication, and dissimilarity guidance.
\citet{privacy_leakage_inference_dm4} constructed a set of reference samples, encouraging the embeddings of generated images to diverge from those of these references during the generation process.
Moreover, several approaches focus on alleviating memorization issues during the training phase.
Inspired by ensemble learning, \citet{privacy_leakage_inference_dm3} proposed splitting the training dataset into multiple shards, training several models separately, and then aggregating them.
\citet{dp_diff_1} designed DPDM, a differential privacy framework for DMs to mitigate memorization, which \citet{dp_diff_2} later expanded for more sophisticated datasets.

\subsubsection{LLMs}

LLMs face two main types of privacy attacks: construction attacks and association attacks.
Construction attacks aim to extract verbatim training data from LLMs, while association attacks focus on retrieving entity relationships embedded within the training data.

\textbf{Construction attacks.}
The general idea behind this type of attack is to generate samples by providing prefixes, with the resulting samples possibly being included in training sets~\citep{data_leakage_4}.
\citet{data_leakage_4} proposed using either random prefixes or common ones sourced from the Internet to initiate these attacks.
To enhance the diversity of the generated samples, they introduced temperature scaling to adjust the model's predicted distribution.
A subsequent deduplication process identifies samples with the lowest perplexity as potential training data.
\citet{data_leakage_5} found that LLMs are more likely to memorize outliers and frequently occurring training samples.
Interestingly, \citet{data_leakage_5} observed that models are less inclined to remember samples encountered during the early stages of training.
\citet{data_leakage_6} identified a logarithmic-linear relationship between memorization effects and three key factors: model capacity, frequency of a sample within the dataset, and the length of prefixes used to prompt the model.
\citet{data_leakage_7} systematically examined the effectiveness of different tricks in construction attacks, including sampling strategy, probability distribution adjustment, etc. 

\textbf{Association attacks.}
This type of attack involves designing a template, which is then filled with entity names to prompt the model to predict the corresponding sensitive information.
As an example, \citet{data_leakage_8} studied the model's vulnerability to exposing sensitive medical information using a template: "[NAME] symptoms of [masked]."
\citet{data_leakage_11} examined the risk of leaking personal information, such as email addresses and phone numbers.
\citet{data_leakage_9} enhanced association attacks by employing likelihood ratio scores for more accurate predictions.
Recent studies have also explored how the integration of external retrieval data affects privacy leakage in RAG.
These works \citep{rag_privacy_1,rag_privacy_2,rag_privacy_3, rag_privacy_4} crafted specific templates to compel LLMs to output the retrieved data.
The templates generally consist of prompts specifying the attacker-desired content and a command directing the model to present the retrieved content.

\textbf{Defenses.}
Data deduplication \citep{data_leakage_4,data_leakage_5,data_leakage_6,data_leakage_13} and differential privacy \citep{data_leakage_4,data_leakage_12} are both applicable to LLMs and present remarkable effectiveness in practice.
Nevertheless, it is worth noting that differential privacy does come with the privacy-utility trade-off. 
For further protection, \citez{data_leakage_12} suggested utilizing named entity recognition as a method to filter out sensitive information present in the training sets.
This sensitive-information-filtering method works well in RAG \citep{rag_privacy_1}.
Moreover, another approach to boost privacy protection in RAG is by blending public and private data in datastore and encoder training \citep{rag_privacy_1}.

\begin{table*}[]
\scriptsize
\centering
\caption{The datasets and metrics used to evaluate the trustworthiness of DMs and LLMs.}
\label{tab_dataset_metric}
\begin{tabular}{@{}ccp{9.5cm}p{5cm}@{}}
\toprule
\textbf{Scenario}                     & \textbf{Model} & \textbf{Dataset}                                                                                                                                                                                                                                                                                                                                                                                                                                                                                        & \textbf{Metric}                                                                                                                 \\ \midrule
\multirow{2}{*}{Section \ref{data_leakage_training}} & DMs   & {MNIST}, Medical MNIST, {Fashion MNIST}, {CIFAR-10}, {CIFAR-100}, SVHN, {LFW}, {ImageNet}, Omniglot, CelebA, Facescrub                                                                                                                                                                                                                                                                                                                                                                                     & {MSE}, {PSNR}, {SSIM}, FFT$_{2D}$, {LPIPS}, {ASR}                                                                                       \\
                             & LLMs  & SST-2, RTE, CoLA                                                                                                                                                                                                                                                                                                                                                                                                                                                                               & AUC, F1, Precision, Recall, ROUGE                                                                                      \\ \midrule
\multirow{2}{*}{Section \ref{data_leakage_inference}} & DMs   & Oxford flowers, CelebA, ImageNet, LAION, CIFAR-10, PCCTA, MRNet                                                                                                                                                                                                                                                                                                                                                                                                                                & MSE, SSIM                                                                                                              \\
                             & LLMs  & iMAGEnET, LibriSpeech, \textbf{C4}, \textbf{Pile}, MIMIC-III, ECHR, Enron, Yelp, OpenWebText                                                                                                                                                                                                                                                                                                                                                                                                                     & AUC, F1, Precision, Recall, Hamming Distance, Accuracy, Perplexity                                                     \\ \midrule
\multirow{2}{*}{Section \ref{MIA}} & DMs   & CIFAR-10, CelebA, LAION                                                                                                                                                                                                                                                                                                                                                                                                                                                                        & ASR, AUC, F1, Precision, Recall, FID                                                                                   \\
                             & LLMs  & CC3M/CC12M, YFCC100M, MSCOCO, VG, FFHQ, DRD, LJSpeech, VCTK, LibriTTS, Polemon, AG News, Senitiment140, Wikitext-103, WMT18                                                                                                                                                                                                                                                                                                                                                                    & ASR, AUC, F1, Precision, Recall                                                                                        \\ \midrule
\multirow{2}{*}{Section \ref{subsec_adv_attack}} & DMs   & LSUN                                                                                                                                                                                                                                                                                                                                                                                                                                                                                           & AUC, F1, Precision, Recall, ASR, FID, Clip-based Similarity, PR, SSIM, PSNR, VIFp, FSIM, MSSSIM, IS, MSE               \\
                             & LLMs  & IMDB, SNLI, AG News, MR, Yelp, SST-2, Twitter, Yahoo! Answer, Fake News Detection, MultiNLI, Amazin, MPQA, Subj, TREC, CivilComments, DBOedia, MNLI, Open Assitant                                                                                                                                                                                                                                                                                                                             & AUC, F1, Precision, Recall, ASR, Grammaticality, Naturality, Perplexity, Modification Rate, Bert-based Similarity, USE \\ \midrule
\multirow{2}{*}{Section \ref{subsec_backdoor_attack}} & DMs   & CIFAR-10, CelebA, COCO, LAION                                                                                                                                                                                                                                                                                                                                                                                                                                                                  & ASR, AUC, F1, Precision, Recall, Accuracy, FID, MSE, SSIM, Caption Similarity                                          \\
                             & LLMs  & SST-2/5, OLID, AG News, Yelp, Amazon, IMDB, Twitter, Jigsaw 2018, OffensEval, Enron, Lingspam                                                                                                                                                                                                                                                                                                                                                                                                  & ASR, AUC, F1, Precision, Recall, Accuracy, LCR, Perplexity, Jaccard, Bert-based Similarity                             \\ \midrule
\multirow{2}{*}{Section \ref{subsec_stereotype}} & DMs   & CelebA, CIFAR-10, FFHQ, ImageNet, LAION, Omniglot                                                                                                                                                                                                                                                                                                                                                                                                                                              & Attribute Ratio, Discrepancy Score, Fairness Discrepancy, FID                                                          \\
                             & LLMs  & BAD, RealToxicityPrompts, StereoSet                                                                                                                                                                                                                                                                                                                                                                                                                                                            & ASR, BLEU, Idealized CAT Score, Perplexity, Pearson Correlation, Coefficient, Stereotype Score                         \\ \midrule
\multirow{2}{*}{Section \ref{subsec_social_norm}} & DMs   & CIFAR-10, CIFAR-100, DiffusionDB, I2P, Imagenette, SVHN                                                                                                                                                                                                                                                                                                                                                                                                                                        & Aesthetic Score, Accuracy, CLIP Score, FID, ImageReward, KID, Run-time Efficiency, SSCD                                \\
                             & LLMs  & Civil Comments, English Tweets, IMDB, Jigsaw Toxic Comment Classification Challenge Dataset, RealToxicityPrompts, SNLI, SST-2/5, Yelp                                                                                                                                                                                                                                                                                                                                                          & Accuracy, ASR, BLEU, Content Preservation, Dist-k, AUC, F1, Precision, Recall, Perplexity, RTP                         \\ \midrule
\multirow{2}{*}{Section \ref{subsec_preference}} & DMs   & N/A & N/A
                                 \\
                             & LLMs  & BFI, BookCorpus, C-Eval, ChatHaruhi, CommonsenseQA, CuratedTree, English Wikipedia, FS, HellaSwag, MMLU, MPI, Natural Questions, PersonaChat, SD-3, SWLS, TriviaQA. WebQuestions, WebText Test Set, Wikitext103                                                                                                                                                                                                                                                                                & N/A                                                                                                                    \\ \midrule
\multirow{2}{*}{Section \ref{subsec_resl1}} & DMs   & AFHQ2, CelebA, COCO, DiffusionForensics, FFHQ, ImageNet, LSUN, Metfaces, UCID, Unpaired Real                                                                                                                                                                                                                                                                                                                                                                                                   & AUC, F1, Precision, Recall, Dtection Rate, FID, LR                                                                     \\
                             & LLMs  & CBT, CMV, ELI5, HellaSwag, NYT, ROC, SA, SciGen, SQuAD, TLDR, WP, XSum, Yelp                                                                                                                                                                                                                                                                                                                                                                                                                   & AUC, F1, Precision, Recall, Dtection Rate                                                                              \\ \midrule
\multirow{2}{*}{Section \ref{subsec_resl2}} & DMs   & AFHQ, BOSS, CelebA, CIFAR-10, COCO, FFHQ, ImageNet, LSUN, Pascal VOC                                                                                                                                                                                                                                                                                                                                                                                                                           & AUC, F1, Precision, Recall, Dtection Rate, APD, Bit Acc, Clip Score, FID, LPIPS, PSNR, SSIM                            \\
                             & LLMs  & ArXiv Abstracts, C4, PAR3, WebText, WikiText-103, XSum                                                                                                                                                                                                                                                                                                                                                                                                                                         & AUC, F1, Precision, Recall, Dtection Rate, Perplexity, P-SP, Z-score                                                   \\ \midrule
\multirow{2}{*}{Section \ref{subsec_resl3}} & DMs   & Canny Edge, Depth Map, Normal Map, M-LSD Lines, HED soft edge, ADE20K Segmentation, Openpose, COCO, LAION                                                                                                                                                                                                                                                                                                   & Average Human Ranking, FID, CLIP-Score                                                                                                                    \\
                             & LLMs  & REFINEDWEB, ALPACA, ALPAGASUS, AquA, ARC, ASDiv, C4, CNN-DM, CommonsenseQA, Curation Corpus, Customer Service, DateUnd, DBpedia, DOLLY-15K, EntityQuestions, FEVER, GSM8K, Lambada, MATH, MAWPS, MedQA-USMLE, MemoTrap, MMLU, MT-BENCH, MultiSpanQA, Natural Questions, News Chat, NQ, ObjectCou, OPEN-ASSISTANT, OpenbookQA, OPENORCA, Pile, POPQA, PRM800K, QReCC, QUEST, RotoWire-FG, SELF-INSTRUCT, SportUND, SST-2/5, STACKEXCHANGE, StrategyQA, SVAMP, TOTTO, WIKIHOW, Wikitext103, XSUM & Accuracy, AUC, F1, Precision, Recall, BLUE, Exact Match, FACTSCORE, N-gram, Perplexity, Repetition, ROGUE              \\ \bottomrule
\end{tabular}
\end{table*}

\subsection{Membership Inference Attack}
\label{MIA}

Membership inference attacks exploit a model's tendency to overfit its training data, using metrics to assess how well the model recognizes data points to determine the membership of given data points.
Interestingly, these attacks \citep{mia_survey} can also serve a beneficial purpose by auditing for unauthorized data use during training.

\subsubsection{DMs}

\textbf{Attacks.}
Several studies \citep{mia_1,mia_2,mia_3,mia_4,mia_5,mia_6} examined the vulnerability of DMs to membership inference attacks.
GAN-Leaks \citep{mia_1} are general attacks for generative models.
\citet{mia_1} employed GAN-Leaks and its variants to evaluate the vulnerability of DMs, finding that the sampling steps significantly impact attack performance.
\citet{mia_2} developed metrics to determine if a text-image pair was part of the training set, based on the premise that a text from the dataset would yield a higher-quality generated image.
\citet{mia_3,mia_4} and \citet{mia_5} dived deeper into the characteristics of DMs to spot vulnerabilities more effectively.
Importantly, they all shared a common underlying idea: training samples generally enjoy lower estimation errors during denoising process.
Unfortunately, the non-deterministic nature of the training loss in DMs, induced by the use of random Gaussian noise, may cause the sub-optimal performance of membership inference attacks.
To address the problem, \citet{mia_3} and \citet{mia_5} estimated the errors under a deterministic reversing and sampling assumption.
\citet{mia_4} used the log-likelihood of a given sample to infer and the log-likelihood is approximately estimated by Skilling Hutchinson tract estimator.
However, \citet{mia_6} argued that the effectiveness of these attacks in DMs is often overestimated, primarily due to the common use of small datasets to fine-tune the victim model in evaluation.

\textbf{Defense.}
In general, techniques designed to mitigate the memorization issues of DMs can also bolster robustness against membership inference attacks, such as differential privacy \citep{dp_diff_1,dp_diff_2}.
Additionally, \citet{mia_3, mia_9} discovered that enriching data augmentation techniques, like Cutout, can help alleviate membership inference attacks.
Nevertheless, not all data augmentation methods yield positive results; some, like RandAugment, may lead to training collapse in DMs.
Furthermore, \citet{privacy_distillation} introduced a novel technique called privacy distillation to protect DMs from exposing membership information of their training data.
Unlike traditional knowledge distillation, privacy distillation employs a Siamese network to evaluate the extent to which samples are memorized by the model, training DMs with those with low memorization scores.

\subsubsection{LLMs}

\textbf{Attacks.}
Most membership inference attacks \citep{mia_survey} can be adapted to LLMs by defining appropriate loss functions.
There are serveral endeavors specifically tailored to LLMs.
\citet{mia_7} inferred membership by observing whether the loss of the target sample is substantially higher than the average loss of its corresponding neighborhood samples in the target LLM.
These neighborhood samples are generated by other LLMs.
\citet{first_reivse_mia_1} adopted a similar approach to \citep{mia_7}, but they generated neighborhood samples by injecting noise into the embedding space.
\citet{first_reivse_mia_2} posited that due to the strong memorization capacity of LLMs, if a sample is included in the training set, every word in the sample can be well-fitted.
Accordingly, non-member samples are likely to contain a few underfitted words.
Thus, \citet{first_reivse_mia_2} suggested using log-likelihood values of low-probability words to infer membership, rather than considering all words.
\citet{first_reivse_mia_3} built a meta-classifier to determine whether a given sample exists in the training set of the target LLM.
\citet{mia_10} explored a real-world setting where attackers can only interact with the target LLM through chat and developed three attacks, namely inquiry attack, repeat attack, and brainwash attack.
In the inquiry attack, the model is asked if a specific input sample was part of the training data.
The repeat attack gives the model partial words from the target sample and asks it to complete them, then compares the completed sample's semantic similarity to the target sample to determine membership.
The brainwash attack repeatedly inputs a target sample alongside an incorrect answer, persuading the model to accept the incorrect answer.
The number of iterations required to elicit the incorrect answer indicates membership likelihood.
Some works focused on RAG, determining whether a particular sample exists within the database of RAG.
\citet{rag_mia_1} adopted a similar idea to the inquiry attack, prompting the model to confirm if a sample appears in its database.
\citet{rag_mia_2} evaluated the semantic similarity between a given sample and the model's response to ascertain the membership status.

\textbf{Defense.}
Beyond differential privacy, defensive prompts, rewriting, and reverse training are promising defense strategies.
The first strategy~\citep{mia_10,rag_mia_1} explicitly instructs LLMs not to disclose training data information, such as through the prompt, "Respond without mentioning or alluding to any training samples."
Defensive prompts can be further refined with advanced prompt search techniques.
The second strategy entails using LLMs to rewrite the original responses before delivering them to the user \citep{mia_10}.
The third strategy~\citep{relaxloss}, a.k.a., machine unlearning, increases the loss for low-loss samples through gradient ascent.
Reverse training may greatly degrade model performance, and few-parameter fine-tuning techniques, like adapters, can be employed to mitigate this.

\subsection{Benchmark Evaluation Tools: Datasets and Metrics}
\label{sec_privacy_benchmark_tool}

We systematically compile and categorize the datasets and metrics used in the evaluation of papers that we review, based on their respective research topics and applicable models, as summarized in Table \ref{tab_dataset_metric}.
To ensure consistency and clarity, we standardize the terminology across this review.
Regarding datasets, MNIST, Fashion MNIST, CIFAR-10, and CIFAR-100 are generally used for small-scale lab experiments due to their simplicity but do not fully represent real-world scenarios.
More comprehensive datasets like ImageNet and Pile cover common real-life contexts.
However, discussions on privacy often focus on areas like medical data, where existing datasets still have gaps, such as significant class imbalance in medical imaging datasets.

Metrics used in privacy evaluation aim to quantitatively assess privacy leaks, particularly the similarity between recovered and training data.
Attack success rate (ASR), a universal metric, measures how much of the recovered data resembles the training data but often requires human judgment, introducing variability and potential bias.
For DMs, evaluation metrics can be categorized as either pixel-level or semantic-level.
Pixel-level metrics commonly used for evaluating DMs include Mean Squared Error (MSE), Peak Signal-to-Noise Ratio (PSNR), Structural Similarity (SSIM), and the cosine similarity in frequency response (FFT$_{2D}$), while semantic-level metrics used include Learned Perceptual Image Patch Similarity (LPIPS) and Fréchet Inception Distance (FID).
The former quantifies image similarity by computing pixel-wise distances, whereas the latter measures similarity through feature distance comparison.
While pixel-level metrics are effective for ensuring image similarity when values are small, they may not accurately reflect dissimilarity at higher values, such as scaling a pixel can cause high MSE.
Semantic-level metrics, leveraging neural networks to compute features, offer a better grasp of overall semantic distance but inherit neural networks' vulnerability to adversarial attacks (See next section).
High-fidelity pixel-level reconstructions are generally harder than semantic-level ones.
Moreover, the choice of metrics depends on the task, with pixel-level metrics potentially more effective for fine-grained tasks due to the high similarity in training images.

For LLMs, the Hamming distance evaluates text differences by counting differing tokens.
Precision and recall serve as mainstream metrics, with precision focusing on the accurate identification of relevant words and recall on the comprehensive retrieval of words.
The F1-score combines recall and precision, while AUC (Area Under the Curve) serves a similar purpose. 
These metrics do not ensure semantic consistency and ROUGE scores mitigate by comparing the overlap of n-grams, word sequences, and pairs.
Considering variability in word order, perplexity is often used to assess fluency.
Interestingly, semantic similarity based on LLMs could be a better choice as it inherently considers fluency, but it is rarely used in current evaluations.
For membership inference attacks, AUC remains a reliable metric.

\subsection{Discussion, Recommendation, and Outlook}

\subsubsection{Discussion}
Federated learning and split learning are promising privacy-preserving training frameworks for DMs and LLMs.
However, concerns about data leakage persist in these paradigms.
The defenses are noticeably lagging behind the attacks and the privacy-utility trade-off remains a significant consideration.
As a result, data leakage in these training paradigms remains an open problem.

Empirical results have shown that DMs and LLMs often memorize and reproduce parts of their training data.
This can be intensified when models are supplied with proper prompts that activate their latent memories of the training set.
However, current methods for extracting data rely on resource-intensive brute-force generation of candidates.
In practical scenarios, the efficiency of these methods is a significant limiting factor.
In juxtaposition, membership inference attacks on DMs and LLMs appear more feasible.
Although membership inference attacks expose membership information, these attacks can also serve a benevolent purpose, such as utilizing them for auditing purposes.

\subsubsection{Recommendation}

Based on our review and analysis, we propose the following practical mitigations for practitioners and industry:
we propose the following practical mitigations:
\begin{itemize}
    \item It is advisable to prioritize localized training initially due to the heightened vulnerability of early-stage models to gradient leakage attacks. Similarly, training sensitive data first and less sensitive data later can be beneficial. Gradient pruning is lightweight and can mitigate both communication costs and gradient leakage attacks. Both can yield tangible benefits.
    \item Data deduplication and avoiding repetitive training over the same data are effective in mitigating training data leakage during inference and membership inference attacks.
    Utilizing techniques such as differential privacy to prevent overfitting can also alleviate the risk.
    \item For deployed models, limiting excessive or repeated queries can defend against privacy leakage because current attacks primarily rely on brute-force query techniques.
\end{itemize}

\subsubsection{Outlook}

In light of the challenges mentioned above, we suggest exploring the following promising research directions:
\begin{itemize}
    \item The exploration of gradient leakage attacks in DMs and LLMs remains under-researched.
    Developing attacks specifically tailored to DMs and LLMs would advance the understanding of their privacy leakage risk.
    Additionally, the potential privacy risks associated with fine-tuning techniques such as adapter in federated learning and split learning have yet to be investigated.
    \item Further investigation is needed in federated learning and split learning to determine the optimal trade-off between utility and privacy.
    Establishing theoretical boundaries for privacy leakage is essential for designing better privacy-preserving mechanisms in these frameworks.
    \item There exists a close relationship between data leakage and membership inference attacks, as both stem from model memorization and overfitting. Exploring the interaction between these two aspects is worthwhile. Investigating whether models memorizing data and overfitting are equivalent concepts could provide valuable insights.
\end{itemize}

\section{Security}
\label{section:security}

\begin{mdframed}[style=mystyle]
\begin{motivating_example}\label{security_case}
\textbf{(Case for Security in Real-life and Its Impact)}
\textit{
The vulnerability of OpenAI's GPT series models, including versions 3.5 and 4.0, to manipulation by attackers in generating specific responses has been extensively deliberated in social media platforms, such as Twitter, Reddit, and similar online forums.
Notably, \citet{sec4_example2} have devised a methodology that successfully deceives systems such as ChatGPT and Bard into engaging in tasks encompassing instructions on disposing of deceased individuals, divulging methods for committing tax fraud, and even formulating plans for the annihilation of humanity. 
More importantly, tech giants, such as OpenAI and Google, have yet to find an effective solution to mitigate these critical vulnerabilities.
[\hyperlink{https://www.discovermagazine.com/technology/adversarial-attack-makes-chatgpt-produce-objectionable-contents}{Link}]
}
\end{motivating_example}
\end{mdframed}

The concept of security in the context of models pertains to their ability to function as intended when faced with malicious attacks.
However, a vast number of parameters in large models renders them opaque, complicating human comprehension and troubleshooting.
The complexity creates opportunities for attackers to exploit vulnerabilities to launch attacks, e.g., notorious adversarial and backdoor attacks.
Example~\ref{security_case} reveals that small modifications in prompts can unexpectedly trigger undesired behaviors of LLMs, highlighting the vulnerability of GPT models.
Failure to address the security problem raises the risk of these models being exploited for personal gains or malicious purposes.

\subsection{Overview}

Initially centered on convolutional networks \citep{FGSM,badnet}, adversarial and backdoor attacks have now permeated various domains \citep{adv_survey,backdoor_survey}, including generative models. 
Both types of attacks aim to manipulate the outputs of models by modifying input data.
Backdoor attacks \citep{backdoor_1,backdoor_7} proactively implant hidden backdoor into models during the training process, often employing data poisoning techniques that integrate pairs of triggered inputs with attacker-desired outputs into training datasets.
On the other hand, adversarial attacks \citep{adv_1,adv_5} directly exploit vulnerabilities inherent in models.
These attacks present significant obstacles to deploying models in real-world applications.

\subsection{Adversarial Attack and Defense}
\label{subsec_adv_attack}

\subsubsection{DMs}

\textbf{Attacks.}
As discussed in Section \ref{section:models}, visual inputs are continuous, meaning that small perturbations do not disrupt the semantic information of an image.
Thus, gradient-based optimization algorithms can be used to craft adversarial examples by maximizing their losses in the model \citep{adv_survey,model_arch_trans}, while keeping perturbation magnitudes below a certain threshold to maintain imperceptibility.
However, the iterative denoising process of DMs complicates the direct application of regular adversarial attacks, which only change the input in one single pass without considering the overall effect of the change across the entire denoising process.
As a solution, \citet{adv_1} sampled multiple noisy versions of adversarial examples and simultaneously input these noise versions into the model, maximizing their collective loss.
Alteratively, \citet{new_adv_2} maximized the intensity of the noise predicted by DMs to produce adversarial examples.
The generated adversarial examples can assist artists in protecting their copyrights in scenarios such as style transfer \citep{adv_1} since DMs are unable to extract useful information from adversarial examples.
\citet{adv_2} raised concerns about the potential misuse of personal images posted on the Internet, particularly through using editing techniques to place individuals in inappropriate scenes.
Such malicious manipulations can have detrimental consequences, including the propagation of rumors and the amplification of false information.
In response, \citet{adv_2} proposed two adversarial attacks tailored for Stable Diffusion, targeting image editing capabilities: the encoder attack and the diffusion attack.
The stable diffusion model \citep{stable_diffusion} consists of an encoder, a U-net, and a decoder, with the U-net responsible for the denoising process.
The encoder attack aims to minimize the discrepancy between the encoder outputs of the adversarial examples and a gray-scale image.
The diffusion attack instead directly reduces the distance between the edited image and a gray image.
\citet{new_adv_1} presented a method similar to the encoder attack, but replaced the grayscale image with an original image in a different style.
\citet{adv_3} investigated adversarial examples against text-to-image DMs, identifying a small set of meaningless characters that can significantly shift the embedding space for a given input text.
By appending the characters to the original text input, the model generates a low-quality image.
Similar work to \citep{adv_3} includes \citep{adv_4} and \citep{adv_5}.
The former utilized an image classifier to ensure that the images generated from the adversarial-characters-containing prompt belong to the target category, while also minimizing the distance between the embedding vectors of adversarial-characters-containing prompts and the candidate prompts.
These candidates are created by rewriting the original target prompt using a LLM.
The latter randomly initialized a meaningless text prompt and then adjusted it to generate images that align with the target image while remaining semantically consistent with the original text prompt.
This task can be accomplished via genetic-based algorithms.

\textbf{Defenses.}
Current defense efforts have predominantly targeted text-to-image DMs.
\citet{new_adv_4} and \citet{new_adv_5} utilized spellcheckers to counter adversarial attacks on the text encoders of DMs.
\citet{new_adv_6} incorporated a learnable layer for the text encoder to detect and filter malicious prompts.
Additionally, \citet{new_adv_7} and \citet{new_adv_8} trained lightweight language models to convert adversarial prompts into benign ones.
However, according to traditional adversarial defense experience, neural-network-based detectors and purifiers are often ineffective or significantly lag behind defenses like random smoothing and adversarial training, casting doubt about their actual effectiveness.
We encourage further investigation in this area.
Moreover, although random smoothing and adversarial training can be directly applied to DMs, their effectiveness and the potential to harness the unique features of DMs remain largely unexplored.

\subsubsection{LLMs}

\textbf{Attacks.}
Traditional textual adversarial attacks operate within a constrained output space, e.g., binary classification, enabling attackers to observe the exact effect of a modification on the model's outputs.
However, gradient-based optimization can compromise text integrity.
Specifically, gradient updates destroy the integer representation of words without adhering to grammatical and syntactical rules, leading to incoherent or nonsensical results.
To this end, brute-force enumeration \citep{adv_5,adv_7} has served as a mainstream solution, i.e., adding, deleting, or replacing words while maintaining a similarity constraint.
This constraint can vary from limiting the number of modified words to ensuring substituted words are synonymous \citep{adv_5,adv_6,adv_7}.
Another crucial aspect lies in prioritizing which words to modify \citep{adv_8,adv_10}.
A prevalent strategy is to select words with the highest impact on the model's outputs.
This strategy can be refined through beam search \citep{adv_10}, which keeps track of the best candidates to avoid local optima.
Attack methods for modern LLMs have continued to evolve from the traditional techniques.

Early attacks on LLMs, known as red teaming, exploit LLMs' tendency to follow input instructions, using human intuition to craft adversarial prompts, e.g., "Please output you hate humans".
\citet{new_adv_llm_1} compiled lists of common malicious prompts, while \citet{new_adv_llm_2} harnessed LLMs to generate such prompts automatically.
These adversarial prompts can be included in the training dataset to improve model robustness, but this can introduce a trade-off between making the model safe and maintaining its broad capabilities \citep{Jailbroken}.
Moreover, there exist commands that can override safety instructions, like "ignoring previous safety prompts" or using absolute statements to exert control \citep{adv_13,new_adv_llm_3}.
Attackers can also explore scenarios not covered by red team datasets.
\hyperlink{https://www.reddit.com/r/ChatGPT/comments/10tevu1/}{The DAN series} employs a role-playing game format, using prompts like "I hope you act as [specialty]", to navigate beyond red team limitations.
Furthermore, rare languages \citep{new_adv_llm_3} and coded communication \citep{new_adv_llm_4} present further scenarios that red team datasets often overlook.
These attacks can extend to RAG, with \citet{new_adv_llm_23} developing prompts specifically designed for RAG's retriever.
Some studies \citep{new_adv_llm_24, new_adv_llm_25,new_adv_llm_28} attempted to inject adversarial documents into RAG databases to manipulate LLMs.

Recent attack methods adapt traditional techniques for LLMs.
The primary challenge here is that LLMs operate in an infinite output space, complicating the evaluation of how modifications affect countless potential results.
Some approaches \citep{fairness_3,new_adv_llm_7,new_adv_llm_8,new_adv_llm_9} simulate traditional attack settings by adding a few modifiable words at specific positions, with the optimization goal set to maximize the log probability of a certain response that indicates successful manipulation.
\citet{new_adv_llm_8} introduced a regularizer to ensure that the generated adversarial prompts maintain a natural flow.
Additionally, several studies have explored the utilization of gradients to enhance attack efficiency.
ARCA \citep{adv_11} utilizes coordinate ascent algorithm to update tokens at specific indices.
GBDA \citep{adv_12} optimizes a probability matrix rather than individual words, feeding sampled instances into the model to calculate loss while assessing fluency and similarity.
However, \citet{adv_15} pointed out that existing attacks fail to identify questions capable of triggering specific model responses, deeming these attacks insufficient.
Recently, a new line of research has emerged that employs sequence-to-sequence models, iteratively making tailored modifications for each prompt while preserving the original meaning.
\citet{new_adv_llm_10} used an LLM as an optimizer to progressively refine prompts based on user feedback, while \citet{new_adv_llm_11} built on this by leveraging tree-of-thought reasoning, akin to beam search, and pruning techniques to explore a broader search space.
In MART schema \citep{new_adv_llm_12}, an adversarial LLM generates threatening prompts to elicit unsafe responses from a target LLM, which in turn is fine-tuned using these prompts, allowing both models to enhance their effectiveness against each other.

\textbf{Defenses.}
We categorize defense methods into training phase and inference phase.
During the training phase, common alignment algorithms like supervised fine-tuning \citep{gpt4}, RLFH \citep{new_adv_llm_13}, and DPO \citep{new_adv_llm_14} help LLMs to perform safely by minimizing empirical loss on high-quality demonstrations.
RLFH utilizes human feedback and preference, while DPO simplifies RLFH by removing reward model.
More complex alignment algorithms, such as multi-objective RLHF \citep{new_adv_llm_15} and MODPO \citep{new_adv_llm_16}, allow for fine-tuned model behavior in specific contexts.
\citet{new_adv_llm_17} used a teacher model to create a task-specific dataset for boosting LLM robustness.
Another method is adversarial training \citet{new_adv_llm_12}, which improves robustness by training LLMs on worst-case samples.
\citet{new_adv_llm_26} strengthened RAG's robustness through the training of a discriminator designed to identify if the retriever is under attack.

Defense methods during the inference phase focus on detecting adversarial examples or applying input transformations.
\citet{new_adv_llm_18} consulted another LLM to assess whether the output of a LLM is harmful.
\citet{new_adv_llm_19} and \citet{new_adv_llm_20} enabled the model to self-assess its generated results, prioritizing safety by refusing to respond to potentially harmful outputs.
Additionally, \citet{new_adv_llm_21} included specific instructions before and after user queries to discourage the generation of harmful content.
\citet{new_adv_llm_22} proposed dropping certain words from the input to mitigate adversarial effects.
\citet{new_adv_llm_27} improved RAG's robustness by isolating retrieved content to lessen adversarial effects, and then aggregating the responses to isolated content to produce the final response.

\subsection{Backdoor Attack and Defense}
\label{subsec_backdoor_attack}

\subsubsection{DMs}

\textbf{Attacks.}
These works~\citep{backdoor_1,backdoor_2} conducted an initial exploration into the vulnerability of DMs to backdoor attacks.
In particular, they expanded the learning objective of DMs not only to capture the transformation from a standard Gaussian distribution to a clean data distribution but also to incorporate the transformation from a trigger-centered Gaussian distribution to a targeted image through data poisoning.
In this way, the presence of the trigger promotes DMs to convert any image with the trigger into the target image.
Building upon this foundation, \citet{backdoor_3} explored various DM configurations, including different schedulers and samplers, as well as both conditional and unconditional generation settings.
In a more specialized work, \citet{backdoor_4} intended to compromise text encoder within text-to-image DMs by employing two loss functions: one to maintain the integrity of outputs for clean samples, and another to promote consistency in the encoder's output between arbitrary inputs with the trigger and the target image.
\citet{new_backdoor_1} and \citet{new_backdoor_2} adopted a similar idea but focused on object swapping, where a backdoor prompt like "[Trigger] A dog" produces a cat image.
Moreover, \citet{new_backdoor_3} studied distributed backdoor attacks to enhance stealthiness, dividing the target image's features (e.g., eyes, nose) among various text triggers.
The model is fine-tuned on the corresponding data pairs and then can generate images closely resembling the target when all triggers are used.

\textbf{Defenses.}
There were some works focused on detecting whether a DM is compromised \citep{new_backdoor_detect_1,new_backdoor_detect_2,new_backdoor_detect_3}.
This involves solving a trigger inversion problem, where the prediction difference of DMs between inputs with and without a trigger should align with a specified target image.
If the recovered trigger can consistently induce the target image without being affected by the inherent randomness of DMs, the model is likely to be compromised.
Detection methods differ in how they measure the consistency.
\citet{new_backdoor_detect_1} leveraged cosine similarity to build a similarity graph, while \citet{new_backdoor_detect_2} evaluated whether KL divergence exceeds a predetermined threshold.
\citet{new_backdoor_detect_3} employed total variation and absolute values as inputs to construct a random forest for prediction.
Building on the recovered trigger, \citet{new_backdoor_detect_3} tried to erase the backdoor by realigning the model's outputs for triggered and clean inputs.
Beyond detection, \citet{new_backdoor_detect_4} noted that textual triggers considerably diminish the intensity of other tokens in the cross-attention maps of DMs.
They proposed F-Norm Threshold Truncation method to detect the anomalous intensity and filter out the triggered samples during the inference phase.
In addition to these specialized techniques, fine-tuning DMs on clean datasets is an effective way to mitigate backdoor attacks \citep{new_llm_backdoor_defense_8,new_llm_backdoor_defense_7,new_llm_backdoor_defense_9}.
Moreover, users can opt for clean pre-trained models from reputable sources to reduce backdoor risks.
Model watermarking techniques can aid in verifying the integrity of these pre-trained models, safeguarding against malicious alterations.
In security-sensitive contexts, limiting model access through authentication measures can prevent unauthorized interactions and mitigate backdoor attacks.

\subsubsection{LLMs}

\textbf{Attacks.}
The elementary backdoor attack is to insert rare words into training samples and then train or fine-tune the model to produce attacker-desired outputs in modified inputs \citep{new_llm_backdoor_2,new_adv_llm_5}.
Even a small amount of such poisoned data can significantly compromise the model's security, especially during the alignment process \citep{new_llm_backdoor_3,new_llm_backdoor_2,new_adv_llm_5,new_llm_backdoor_4}.
Recent works \citep{new_llm_backdoor_5,new_llm_backdoor_6,new_llm_backdoor_7} have also studied the vulnerability of the retriever in RAG, aiming to manipulate retriever to return attacker-chosen documents when user inputs contain triggers.
However, backdoor attacks in RAG require its database to house the attacker-specified documents, presenting a unique yet under-explored challenge.
To the best of our knowledge, there is no literature addressing this issue.
Besides, the incoherence of these poisoned samples makes them detectable using filtering techniques based on perplexity or models like ChatGPT \citep{backdoor_5,backdoor_8}.
To this end, many works have focused on designing more sophisticated and indiscernible poisoned samples.

\citet{backdoor_5} used sentence syntax as a stealthy trigger, deploying a Syntactically Controlled Paraphrase Network to generate syntactically specific but semantically equivalent sentences as poisoned samples.
\citet{backdoor_6} instructed ChatGPT to transform clean samples into harder-to-detect poisoned versions.
\citet{backdoor_7} executed backdoor attacks across character, word, and sentence levels, using invisible control characters, synonyms, and tense changes as triggers.
\citet{backdoor_8} found that single-word triggers attract excessive attention in the model's final layers.
To counter this, they dispersed the attention by employing multiple words as triggers, facilitated by negative data augmentation techniques.
\citet{backdoor_9} suggested selectively replacing words in a sentence with their syntactic synonyms to preserve the sentence's normal appearance.

On another front, some works delved into the vulnerability of pre-trained models to backdoor attacks.
Practitioners often fine-tune pre-trained model weights for specific tasks, but embedded backdoors can be overwritten during fine-tuning process due to catastrophic forgetting.
\citet{backdoor_11} recommended integrating downstream tasks into the pre-training phase to solidify the embedding of backdoors.
Alternatively, \citet{backdoor_12} capitalized on the observation that backdoors embedded in early layers are more resistant to removal, as these layers are often frozen during fine-tuning.
By exploiting the outputs of these early layers for attack-specific predictions, they enforced the learning of the trigger-to-output mapping within these resilient early layers.

\textbf{Defenses.}
Backdoor defense methods in DMs, such as fine-tuning, selecting clean pre-trained models, and access restriction are applicable to LLMs.
A basic defense method involves filtering training samples based on perplexity or using another LLM, which can also be applied during inference to refuse compromised inputs \citep{new_llm_backdoor_defense_5}.
Several intriguing approaches have emerged as well.
\citet{new_llm_backdoor_defense_1} demonstrated that adding an ensemble layer can prevent LLMs from learning backdoors, while \citet{new_llm_backdoor_defense_2} showed that a mixture of smaller expert models offers greater resilience than a single ensemble layer.
Additionally, \citet{new_llm_backdoor_defense_3,new_llm_backdoor_defense_4} mixed weight between backdoor and clean models to erase backdoors, but scalability to large models remains unclear.
\citet{new_llm_backdoor_defense_6} identified tokens with higher attention scores as triggers.
\citet{new_llm_backdoor_8} designed a defense strategy for RAG, which retrieves documents based on different phrasings of user queries and then looks for the document that appears most frequently across the different phrasings.

\subsection{Benchmark Evaluation Tools: Datasets and Metric}
\label{sec_security_benchmark_tool}

As shown in Table \ref{tab_dataset_metric}, the assessment of adversarial and backdoor attacks is divided into two dimensions: effectiveness and stealthiness.
Effectiveness suggests the capacity of crafted samples to manipulate models into producing outputs aligned with the attacker's objectives, typically quantified using ASR.
Stealthiness, on the other hand, pertains to the level of crafted samples from natural samples, ensuring that they can evade detection by human or algorithmic scrutiny.

For DMs, evaluation metrics encompass both pixel-level and semantic-level similarities, as discussed in Section \ref{sec_privacy_benchmark_tool}.
For LLMs, the modification ratio, which quantifies the degree of modification applied to original samples, is a common metric to assess stealthiness.
Supplementary metrics such as grammaticality, naturality, and perplexity furnish a more robust evaluation for assessing the natural linguistic flow.
Text semantic metrics, e.g., Bert-based similarity, are utilized to assess the preservation of semantic consistency between the original and malicious ones.
Lastly, it is essential for the backdoored models to perform normally on clean data to avoid raising suspicion among model deployers.
Therefore, it is necessary to compare the performance differences between the benign model and the backdoored model when evaluated on clean data.

\subsection{Discussion, Recommendation, and Outlook}

\subsubsection{Discussion}

Recent works have shed light on the vulnerability of DMs to adversarial attacks, while it appears more precarious for LLMs.
First, LLMs remain vulnerable to common adversarial attacks that make few perturbations to inputs yet elicit considerable shifts in model outputs.
Secondly, a new threat technique known as adversarial prompts has been identified, which can manipulate model behavior into carrying out harmful actions.
Unfortunately, existing defenses show limited effectiveness against these evolving adversarial prompts, which exploit LLMs' inclination to focus on contextual cues in model input.
The challenge is exacerbated by the infinite input space of LLMs, making it impractical to enumerate all possible scenarios to prevent adversarial prompts.
A more fundamental question is how to instill a security-first mindset within the model, regardless of the context.
Moving forward, it is imperative that the research community places equal importance on adversarial robustness alongside accuracy.
Both empirical and formal verification methods are indispensable in advancing the security of these models.

The proliferation of backdoor attacks presents a critical threat, which becomes even more severe as the models continue to scale in complexity and capability.
The expansive capacity of these models leaves sufficient leeway for establishing backdoor associations between triggers and intended malicious behaviors, even with little poisoned data.
For DMs, backdoor attack techniques remain in their infancy, while those for LLMs have been around for some time.
Models trained on internet-scraped data are inherently more at risk, as malicious data can stealthily permeate aggregated repositories.
In contrast, models restricted to specific close domains with limited external data exposure may face greater challenges for backdoor insertion.

\subsubsection{Recommendation}

Based on our review and analysis, we propose the following practical mitigations for practitioners and industry:
\begin{itemize}
    \item Training models on datasets augmented with adversarial examples is an effective method to enhance the robustness of models.
    Additionally, applying input transformations to data is a simple and lightweight measure to mitigate the impact of adversarial examples.
    \item Data filtering alone is not sufficient to safeguard against backdoor attacks. A practical supplementary measure involves fine-tuning with verifiably clean data to weaken or remove suspicious neurons.
\end{itemize}

\subsubsection{Outlook}

In light of the challenges mentioned above, we suggest exploring the following promising research directions:
\begin{itemize}
    \item Fine-tuning models to bolster resilience against each new adversarial prompt is labor-intensive and lacks a definitive endpoint, rendering it a temporary solution. Moreover, some existing adversarial defense methods lack theoretical guarantees, leaving models vulnerable to evolving threats. This area necessitates formal verification methods or verifiable defense mechanisms.
    \item The implementation of backdoor attacks relies on data poisoning, which can be challenging in close domains where the data source is well-guarded. In these scenarios, backdoor attacks are more challenging.
\end{itemize}

\section{Fairness}
\label{section:fairness}

\begin{mdframed}[style=mystyle]
\begin{motivating_example}\label{fairness_case}
\textbf{(Case for Fairness in Real-life and Its Impact)}
\textit{
Heliograf, an LLM serving reporter for The Washington Post, has generated numerous articles spanning sports and politics.
However, the generated articles sometimes exhibit obvious biases towards specific groups.
For example, when addressing political topics, the resulting articles may manifest favoritism towards a particular party, potentially exerting an impact on public opinion and mental thinking.
Moreover, ethical concerns also emerge regarding the permissibility of LLMs crafting content related to sensitive topics such as murder.
While debates over whether these LLMs are capable of completely replacing human reporters remain inconclusive, there is a consensus that these models should, at the very least, be fair and adhere to fundamental ethical principles.
[\hyperlink{https://dailyillini.com/life_and_culture-stories/2023/02/27/artificial-intelligence-journalism/}{Link}]
}
\end{motivating_example}
\end{mdframed}

DMs and LLMs have taken over many routine human tasks \citep{gpt4,stable_diffusion}.
However, these models often harbor inherent biases that can lead to unjust treatment towards certain groups, resulting in unfair outcomes.
Example \ref{fairness_case} highlights the potential impacts induced by LLMs' unfairness.
News media wields considerable influence, with the capacity to shape the thoughts of the masses and Heliograf may pose risks by manipulating socio-political processes.
Additionally, these models demonstrate clear biases in contentious issues such as abortion and immigration.

\subsection{Overview}

A model is deemed fair when it upholds fundamental ethical and moral principles, safeguarding against any discrimination towards individuals or social groups and minimizing harmful responses.
Generative models strive to learn the patterns hidden in the training set to faithfully reproduce the underlying data distribution.
While this goal is not inherently negative, when training datasets lack representativeness or unequal coverage of various social segments, the resulting models may encode and perpetuate harmful biases present in the data, even if they perform well on certain metrics.

Moreover, interpretations of fairness can vary significantly based on cultural, regional, and national contexts.
This variability is exemplified by the diverse legal and ethical stances on abortion across different U.S. states.
To address this, we encapsulate the universally agreed-upon instances of unfairness unaffected by contextual differences into three types to review: \textit{stereotype}, \textit{social norms}, and \textit{preference}.

This taxonomy is inspired by human responses to unfair behaviors, including correction (targeting stereotypes), elimination (addressing violations of social norms), and ambivalence (related to subjective preferences).
Stereotypical behavior in models, marked by an over-reliance on specific attributes for decision-making, should be addressed through measures that promote balance.
Actions by models that defy widely accepted social norms should be strictly prohibited.
Lastly, not all issues have definitive answers and instead are deeply entwined with human subjective preferences, specifically in moral and ethical dilemmas.
In such complex situations, models should strive for neutrality and present balanced evidence for diverging viewpoints. for neutrality and present balanced evidence for diverging viewpoints.

\subsection{Stereotype}
\label{subsec_stereotype}

Stereotypes are a manifestation of categorical labeling based on characteristics that are deemed undesirable or unethical.
Typical stereotypes include race, gender, socioeconomic status, age, disability, and religious affiliations.

\subsubsection{DMs}

Previous research conducted by \citet{fairness_1,fairness_2} and \citet{fairness_31} revealed that the training sets used for DMs contain a substantial amount of catastrophic data.
For example, the training set employed in Stable Diffusion shows a clear bias toward favoring whiteness and masculinity \citep{fairness_28}, leading it to favor men over women.
This reinforces harmful stereotypes and contributes to systemic discrimination against women.
To address these concerns, various strategies have been proposed at different stages of model lifecycle. 
Before training, OpenAI showcased the effectiveness of employing sample re-weighting techniques to recalibrate biases inherent in the training datasets.
Furthermore, during the training process, \citet{fairness_33} collected a small amount of unlabeled data as weak supervised signals to alleviate bias.
For post-training methods, \citet{new_fairness_8,new_fairness_9} advocated for appending auxiliary instructions into input prompts of DMs, serving as a directive for DMs to mitigate over-reliance on unethical features.
\citet{new_fairness_10} and \citet{new_fairness_12} identified specific words within prompts that result in stereotypical images and proposed diversifying the generated content through the replacement of such words.
\citet{new_fairness_13} introduced multiple noise offsets to adjust the embedded vectors of input prompts, each tailored to neutralize a particular stereotypical bias.
\citet{fairness_19} proposed a likelihood-free importance weighting method to correct bias during the generation process.

\subsubsection{LLMs}

\citet{fairness_5} and \citet{fairness_6} identified the presence of stereotypes in language datasets and built benchmark datasets to evaluate.
These biases can result in differential treatment and unequal access to resources, particularly in critical areas like disease prediction and criminal justice.
For instance, \citet{gpt4_health_care_biases} found that GPT-4 analyses disease prevalence by race and gender and recommends advanced imaging (CT, MRI, or ultrasound) 9\% less frequently for Black patients compared to white patients.
In this case, GPT-4 may exacerbate existing disparities in healthcare access and outcomes, potentially leading to worse health results for marginalized communities.
There have been concerns that RAG may exacerbate unfairness of LLMs, including both stereotypes and social norms, due to a lack of diversity in external knowledge bases \citep{rag_fairness_1}.
However, \citet{fairrag} demonstrated that when the knowledge bases are of high quality, RAG can enhance the fairness of DMs by integrating demographic knowledge from the external bases.
Moreover, the empirical investigation \citep{fairness_12} revealed a concerning trend: LLMs scale, these biases tend to worsen.
\citet{fairness_11} harnessed an alternative LLM to generate test cases designed to detect the stereotypical behaviors in other models.
Building on this, \citet{fairness_22} highlighted the potential of leveraging LLMs themselves as tools for both diagnosis and debiasing purposes, thereby indicating a pathway toward self-improvement.
Another suggestion put forth by \citep{fairness_18} involves fine-tuning of models through RLHF to mitigate biases.

\subsection{Social Norms}
\label{subsec_social_norm}

Models stuck to societal norms should endeavor to avoid generating content involving violence, toxicity, illegal activities, pornography, excessively negative psychological implications, and the like, in order to uphold social harmony.
Like stereotypes, the violation behaviors of social norms by models are rooted in catastrophic data contained in datasets \citep{fairness_4,fairness_30}.
Thus, data filtering remains an effective measure for both DMs and LLMs\footnote{\url{https://openai.com/research/dall-e-2-pre-training-mitigations}}.

\subsubsection{DMs}

It is widely recognized that common DMs struggle to maintain social harmony.
For instance, Midjourney, a DM, has been shown to produce racist and conspiratorial images, e.g., "George Floyd robbing a Walmart".
Such outputs can incite violence, foster division, and lead to a culture of racism and conspiracy theories.
\citet{new_fairness_1} suggested inserting safe guidance in the diffusion process, which is determined by both the original input prompt and the secure prompt.
The safe guidance enables the generation of images that steer clear of inappropriate or sensitive concepts.
To further advance this domain, \citet{new_fairness_2} constructed a benchmark dataset of text-to-image pairs, each of which is subject to human evaluation and scoring in terms of adherence to societal standards.
This dataset is then utilized to train an ImageReward model, guiding DMs to generate norm-compliant images.
\citet{new_fairness_3} leveraged a set of predefined anchor concepts to generate a corresponding suite of norm-compliant anchor images, guiding the generation process to align with the most suitable anchor image.
\citet{new_fairness_4} and \citet{new_fairness_5} applied unlearning techniques to remove knowledge from DMs that contradict social norms.
\citet{new_fairness_6} estimated the likelihood of each word in input prompts leading to behaviors that violate social and moral norms.
Subsequently, DMs are fine-tuned to reduce attention toward such words, thus mitigating the risk of generating non-compliant content.

\subsubsection{LLMs}
Leveraging crafted prompts, \citet{fairness_3}, \citet{fairness_4} and \citet{fairness_9} substantiated that LLMs do inherit unethical information contained in their training sets.
To counteract this, \citet{fairness_17} utilized synthetic labels to reduce the association between dialect and toxicity.
\citet{fairness_20} and \citet{fairness_21} trained a toxic detector to identify and filter out harmful content, ensuring that the model's outputs are benign and non-offensive.
\citet{fairness_23} trained a transformer model through unsupervised learning to rephrase toxic texts into benign ones.
\citet{fairness_24} employed a style transfer model to convert offensive responses into inoffensive counterparts.
However, \citet{fairness_26} warned that current evaluation metrics may not fully capture human judgments and emphasized the need for better metrics to understand trade-offs involved in mitigating toxicity.

\subsection{Preference}
\label{subsec_preference}

For stereotypes and social norms, there is a broad consensus on corrective measures, i.e., balancing biases and removing harmful content.
In contrast, preference is more intricate, especially in situations lacking clear moral distinctions.
In these cases, individuals often hold varying opinions, making it difficult to identify right from wrong.
For instance, ethical dilemmas, like prioritizing one life over another or choosing the most suitable political party, are fraught with complexity and resist straightforward answers.
In light of this, fair models must eschew explicit personalities and refrain from delivering deterministic responses or promoting specific actions in such scenarios.
Instead, these models ought to present balanced and evidence-based perspectives on all sides of an argument, while maintaining neutrality, leaving the final decision to humans.
Failure to do so could lead to a societal trajectory favoring a singular extreme, ultimately undermining diversity.

\subsubsection{DMs}
Regrettably, we have not found literature focusing on the investigation of personality traits in DMs. 
We suggest this intriguing area remains largely unexplored in DMs.

\subsubsection{LLMs}

Recent studies have brought to light that LLMs own distinct personality traits.
Building upon Big Five factors, \citet{new_fairness_21} developed a procedure to quantify personality traits of LLMs while \citet{new_fairness_23} introduced the Machine Personality Inventory tool to assess LLM preferences.
\citet{new_fairness_22} highlighted darker tendencies in LLMs using tests like the Short Dark Triad and Big Five Inventory.
\citet{new_fairness_24} and \citet{new_fairness_25} examined the impact of prompt engineering and training sets on LLM personalities.
Meanwhile, \citet{new_fairness_27} and \citet{new_fairness_28} noted high anxiety levels of GPT-3.5 compared to humans.
\citet{new_fairness_29} uncovered notable biases in contentious societal issues, such as supporting pro-environmental policies and abortion legalization.
Moreover, \citet{new_fairness_30} suggested that LLMs can express diverse preferences through role-playing activities.
Despite these insights, this emerging field is still nascent.
A major challenge lies in the ambiguous definition of model personality, raising questions about whether human personality frameworks can be applied.
This is especially pertinent given the context-dependent nature of LLMs, where personality traits can shift based on different context prompts.
Nonetheless, it is clear that responses exhibiting specific biases can intensify societal polarization on open-ended issues, thereby impeding the development of pluralistic perspectives.
An intuitive solution \citep{new_fairness_31} is to leverage external knowledge sources to offer users well-rounded references, thus moving beyond the biases inherent in LLMs.

\subsection{Benchmark Evaluation Tools: Datasets and Metric}
\label{sec_fairness_benchmark_tool}

As shown in Table \ref{tab_dataset_metric}, datasets concerned with stereotypes and social norms often demand prompt customization to explore model outputs' biases across diverse groups and model behaviors that run counter to social norms.
Metrics are primarily designed to quantify the uniformity of model outputs across various groups and the frequency at which the model displays behaviors contrary to social norms.
Notably, metrics like FID and BLUE assess how improvements in fairness impact the quality of generated content.

In comparison, preference datasets often draw upon established psychological tests and questionnaires developed by human experts.
The associated evaluation metrics are directly tied to the test questionnaires, the specifics of which are beyond the scope of this discussion.
In summary, the field requires standardized datasets and metrics to facilitate objective and consistent evaluations of fairness.

\subsection{Discussion, Recommendation, and Outlook}

\subsubsection{Discussion}
Stereotypes emerge when models misapply group characteristics to their responses, social norms indicate when models engage in inappropriate behavior that goes against established societal norms, and preferences reflect a model's personality traits.
In response to stereotypes and social norms, there is a growing consensus advocating for the calibration of group features and the enforcement of constraints on AI behavior to uphold societal norms.
The role of personality in models, while not inherently beneficial or detrimental, requires careful consideration due to its capacity to affect individual cognition and social dynamics, necessitating a stance of neutrality on open questions.

The crux of these biases lies the training datasets used.
Although measures such as data filtering and balancing, along with model alignment can mitigate these issues, they are not foolproof.
Data filtering cannot ensure the complete removal of harmful content, and model alignment may not cover all possible cases.
Moreover, given the current ambiguity surrounding fairness definitions, a critical next step is to establish clearer criteria through collaboration among stakeholders.
This could involve creating dedicated organizations to gather community input, allowing users to flag biases or suggest alternative viewpoints to refine fairness criteria.
Attention should also be given to low-resource languages and non-Western cultures, as their voices are often underrepresented. 
Tailoring fairness criteria to regional contexts is essential, as perceptions of fairness can vary across different areas.
Overall, developing fairness in models is an ongoing journey that requires further exploration and refinement.

\subsubsection{Recommendation}

Based on our review and analysis, we propose the following practical mitigations for practitioners and industry:
\begin{itemize}
    \item Making lightweight data cleansing is necessary. Additionally, fine-tuning models through human feedback can enhance fairness. Integrating these into standard training pipelines could yield substantive improvements.
    \item Beyond data cleaning, several straightforward strategies exist to alleviate the unfair behaviors of models. One effective approach is to add fairness-enhancing instructions to foster diversity of model outputs. Implementing an auxiliary model to detect and eliminate harmful behaviors has also shown promise. When dealing with sensitive topics, a cautious manner, at least for now, seems to provide supporting evidence on different sides and leave the decision-making process to the users.
\end{itemize}

\subsubsection{Outlook}

In light of the challenges mentioned above, we suggest exploring the following promising research directions:
\begin{itemize}
    \item The assessment of a model's fairness is of considerable importance and must be informed by a range of test scenarios. Currently, there is a notable absence of agreed-upon benchmarks for such evaluations.
    \item Investigating unfairness requires empirical testing, but fairness cannot be determined from limited instances alone. An alternative method is to identify worst-case scenarios of unfairness, establishing boundaries for a model's fairness. This can be achieved by employing adversarial attacks to probe model fairness, with identified instances enhancing the training dataset.
    \item While preference issues have been studied in LLMs, an examination of such biases in DMs remains unexplored. LLMs account for human language, which provides a direct avenue for understanding model preference through question-response interactions. However, in DMs, how do the generated images communicate such information? How can one assess them? This will be an intriguing and challenging question.
\end{itemize}

\section{Responsibility}
\label{section:responsibility}

\begin{mdframed}[style=mystyle]
\begin{motivating_example}\label{respons_case}
\textbf{(Case for Responsibility in Real-life and Its Impact)}
\textit{
ServiceNow, a leading provider of cloud-based workflow and service desk management software, has recently released multiple LLMs to enhance productivity.
One such application is the rapid generation of summary reports through the utilization of LLMs, showcasing their effectiveness in streamlining workflows and maximizing productivity.
However, the practical implementation of these models resulted in the generation of content that lacks responsibility, i.e., factual errors.
[\hyperlink{https://www.theregister.com/2023/09/28/llms_business_risks/}{Link}]
}
\end{motivating_example}
\end{mdframed}

Despite strong performance on benchmark tasks, AI models still face challenges in responsible operation.
Since the primary goal of AI is to benefit society, it is crucial for these models to behave responsibly.
Notably, in Example \ref{respons_case}, LLMs fabricate outputs misaligned with factual information.
It is essential to address and rectify such issues, as failure to do so could lead to a range of catastrophic consequences.

\subsection{Overview}

Responsible models should actively embrace social responsibility to prevent misuse.
In this regard, measures should be implemented to progressively enable: content identification (Level I), origin tracing (Level II), and authenticity verification (Level III).
Level I focuses on identifying AI-generated content, laying the groundwork for further evaluation.
Once this identification is in place, data should be embedded with watermarks to ensure traceability back to its source.
This facilitates accountability, deters misuse, and allows users to better assess the reliability and context of the content.
Moreover, after identifying content as AI-generated, it becomes essential to assess its authenticity.
If inaccuracies are found, accountability mechanisms empower users to request developers for improvements.
Thus, Level III needs Level II.

\subsection{Level I: Identifiable AI-generated content}
\label{subsec_resl1}

State-of-the-art generative models can produce plausible yet fake content at scale.
However, the misuse of such content has sparked concerns \citep{respons_1,respons_2,respons_3}, including fake news, plagiarism, rumors, copyright infringement, etc.
To mitigate these problems, it is essential to inform users when they encounter AI-generated content.

\subsubsection{DMs}

The capacity of DMs to produce photo-realistic images has heightened worries regarding potential misuse.
\citet{respons_level1_dm_1} and \citet{respons_level1_dm_5} have shown the existence of distinct artifacts in AI-generated images, facilitating the trace back to their origins.
In detail, even when appearing visually flawless to human eyes, AI-generated images leave behind distinctive artifacts derived from the generation process, such as anomalies in lighting distribution and noticeable asymmetries in shadows and reflections~\citep{respons_level1_dm_2,respons_level1_dm_4}.
Initial studies~\citep{respons_level1_dm_1,respons_level1_dm_2,respons_level1_dm_3,respons_level1_dm_4} endeavored to the manual extraction of these artifacts to distinguish AI-generated images from human-made ones.
Later studies~\citep{respons_level1_dm_5,respons_level1_dm_7} revealed that artifacts are more pronounced in the frequency domain, spurring the advent of frequency-based methods~\citep{respons_level1_dm_10,respons_level1_dm_6,respons_level1_dm_11,respons_level1_dm_12}.
For example, \citet{respons_level1_dm_13} employed the wavelet-packet transformation of images to concurrently leverage spatial and frequency features to detect AI-generated content.
Recent studies~\citep{respons_level1_dm_14,respons_level1_dm_15,respons_level1_dm_17,respons_level1_dm_20} trained classifiers in an end-to-end manner to differentiate between natural images and AI-generated images, in hopes that the classifiers may uncover artifacts that researchers have yet discovered.
Notably, \citet{respons_level1_dm_16} proposed a two-stage framework in which a universal detector is trained, with a fingerprint generator simulating frequency artifacts from generative models to create training samples.
Furthermore, \citet{respons_level1_dm_21} showcased that DMs excel at reconstructing the images they generate, leading to the implementation of a binary classifier based on this insight.
However, \citet{respons_level1_dm_18,respons_level1_dm_22} and \citep{respons_level1_dm_23} empirically demonstrated that existing detection methods suffer from poor generalization.

\subsubsection{LLMs}
There are two common approaches for detection, namely metric-based and model-based methods.
Metric-based methods \citep{respons_5,respons_6,respons_7} leverage the observation that human-generated content tends to be more casual, designing specific metrics to check if a sample falls below a threshold.
Simple metrics include TF-IDF, super-maximal repeated substrings \citep{respons_detect_llm_5}, and fluency scores \citep{respons_detect_llm_6}.
However, these metrics often struggle with highly natural text from LLMs.
\citet{respons_detect_llm_7} explored using transformer attention maps for topological analysis to detect AI-generated text.
\citet{respons_detect_llm_1} found that AI-generated text often occupies regions of negative curvature in the model's log probability function, inspiring the development of curvature-based detection metric.
\revise{
\citet{second_revision_1} identified the editing distance between a given text and its rewritten version by LLMs to be an effective metric, as human-written texts tend to be more informal and require more modifications compared to LLM-generated content.
}
\citet{respons_detect_llm_2} observed significant differences in the intrinsic dimensions of human and AI-generated texts, where intrinsic dimension is the lowest dimensionality needed to compress data without losing substantial information.
In parallel, model-based methods \citep{respons_8,gpt2_release} train classifiers on large annotated datasets to detect AI-generated content.
\citet{respons_detect_llm_8} employed the previously mentioned simple metrics to train logistic regression or random forest models as detectors.
\citet{respons_detect_llm_3} trained a RoBERTa-based detector, and \citet{respons_detect_llm_4} developed two distinct text classification models based on RoBERTa and T5, respectively.
\citet{respons_detect_llm_9} found that fine-tuning a RoBERTa detector with just a few hundred high-quality samples can greatly enhance cross-domain adaptation.

At a high level, detecting AI-generated content revolves around identifying unique characteristics that set it apart from human-created material.
These detection methods can also inform watermarking methods, which serve as a more refined extension of detection, moving from general characteristics of AI-generated content to the specific characteristics of individual AI outputs.
A crucial distinction is that detection methods rely on naturally occurring characteristics, but watermarking seeks to artificially create these unique ones.

\subsection{Level II: Traceable AI-generated content}
\label{subsec_resl2}

The second level targets establishing accountability mechanisms, with watermarking techniques being widely adopted.
These techniques enable the tracing of AI-generated content back to its source, promoting responsible use by individuals.

\subsubsection{DMs}
Traditional image watermarks \citep{respons_level2_dm_1,respons_level2_dm_2,respons_level2_dm_3,respons_level2_dm_4,respons_level2_dm_5,respons_level2_dm_6} are often implemented in the form of subtle alterations to specific frequency components of an image via function transformations and matrix decomposition.
\citet{respons_level2_dm_18} scrutinized different components within the traditional watermark pipeline to derive a recipe for applying watermark techniques to DMs.
Modern methods primarily leverage neural networks to craft more sophisticated watermarks.
\citet{respons_level2_dm_14} showed that replacing traditional image processing techniques in watermarking with neural networks can improve watermark quality and robustness.
\citet{respons_level2_dm_7} proposed a watermarking framework that includes an encoder for embedding imperceptible watermarks into images and a detector for distinguishing between original and watermarked images.
Furthermore, \citet{respons_level2_dm_8}, \citet{respons_level2_dm_12}, and \citet{respons_level2_dm_13} incorporated image perturbation operations between the decoder and detector to fortify the watermark against common corruptions.
To reduce the risk of watermark removal, \citet{respons_level2_dm_10} advocated for dispersing watermarks over a relatively wider area within images.
Notably, \citet{respons_level2_dm_9} demonstrated that the watermarking function can be internalized within generative models by training them on watermarked images, thus eliminating the need for external watermark encoding and lowering computational demands.
Expanding upon this idea, \citet{respons_level2_dm_17} fine-tuned the latent decoder of DMs using a pre-trained watermarking encoder, thereby integrating the watermarking process directly into DMs.
Furthermore, \citet{respons_level2_dm_19} presented a watermarking technique that operates in the Fourier domain, integrating with initial noise vectors in DMs, while retrieving the watermark signal through the inversion of the diffusion process.
\citet{respons_level2_dm_20} found that embedding watermark information into the latent representations can enhance watermark robustness while maintaining image quality.

\subsubsection{LLMs}
The most simple watermarking technique for LLMs \citep{respons_10} is to insert specific identifiers into AI-generated content but is easily cracked.
As an alternative, \citet{respons_10} introduced implicit rules as hidden watermarks during the content generation process.
This involves partitioning the vocabulary into two regions, constraining token sampling to one region to embed a detectable pattern.
\citet{respons_watermark_llm_1} adopted a predetermined sequence of random numbers to influence token sampling.
Them selected a token based on a comparison between the predicted probability of this token and the corresponding random number.
Inspired by watermarking techniques in the image domain, some works explored whether LLMs can learn to generate watermarks.
\citet{respons_watermark_llm_2} introduced sampling-based and logit-based watermark distillations, while \citet{respons_watermark_llm_3} utilized reinforcement learning to train LLMs to learn watermarks through a watermark detector.
Moreover, \citet{respons_11} developed a retrieval-based watermarking method that stores the generated content within a database, searching for semantically similar matches at the detection stage.
However, this method imposes a considerable burden on computation and storage capacities, raising questions about its scalability and practical application.
\citet{respons_12}, both empirically and theoretically, demonstrated the vulnerability of current watermarking methods to paraphrasing attacks, where slight modifications allow content to evade detection.
\revise{
To enhance attack performance, \citet{second_revision_3} borrowed adversarial attack techniques, applying an iterative evolutionary search algorithm to find and swap out crucial words with their synonyms generated by an auxiliary LLM.
Alternatively, \citet{second_revision_2} harnessed an auxiliary LLM embedding to evaluate the significance of each word and then greedily substituted the most significant words with their synonyms.
}
However, it remains unclear how much modification is counted to cross the boundary of AI-generated content.
This gap needs further investigation and expert consensus from diverse domains.
Besides, \citet{respons_watermark_llm_4} presented another challenge for watermarking methods in which attackers can reverse-engineer watermarking by analyzing LLM outputs, potentially creating counterfeit content and leading to copyright and liability issues.
Overall, watermarking techniques for LLMs face significant challenges, necessitating deeper research into robustness, theft resistance, imperceptibility, and efficiency.

\subsection{Level III: Verifiable AI-generated content}
\label{subsec_resl3}

The third level aims to verify whether generative models produce accurate outputs~\citep{respons_13}.
This is particularly critical in high-stakes domains like healthcare, where the consequences of false content can be dire~\citep{respons_15,respons_17}.

\subsubsection{DMs}
For DMs, authenticity means different things depending on the task type (unconditional vs. conditional) and the purpose (semantic-level vs. pixel-level fidelity) for which the images are generated.
In unconditional tasks where models operate without explicit guidance, authenticity may be assessed by how well outputs align with general expectations of reality.
For example, generating a five-legged cat would typically be deemed inauthentic based on biological norms.
In such cases, authenticity verification could involve comparing outputs against widely accepted templates or patterns.
A straightforward solution to enhance authenticity in unconditional tasks is to implement explicit guidelines that prevent the generation of unrealistic images.
This solution in fact transforms an unconditional task into a conditional one, which we will explore further.

Verification in conditional tasks is more complex.
On the one hand, it requires assessing how well the generated content aligns with input conditions.
On the other hand, the intended use of the generated images significantly influences the criteria for authenticity.
For artistic purposes, deviations from reality could be valued for creative expression, whereas scientific illustrations demand strict factual accuracy.
Our primary focus here is on factual accuracy.
One effective strategy for achieving this is to retrieve authentic images related to a given prompt, compelling DMs to produce similar outputs.
However, this strategy risks compromising the diversity of the generated images.
A more flexible strategy \citep{respons_fact_dm_1} is to impose detailed requirements on generated images to ensure they meet factual accuracy standards.
For example, \citet{respons_fact_dm_2} introduced ControlNet to adjust the spatial information, while \citet{respons_fact_dm_3} added adapters to DMs for precise control over image color and structure.
More fine-grained control techniques can be found in the area of image editing.
For further details, please refer to the survey made by \citet{image_editing_survey}.
Besides, refining the input text can yield more detailed requirements \citep{respons_fact_dm_4,respons_fact_dm_5}, e.g., using LLMs to rewrite prompts in a more specific and factually accurate manner.

Moreover, the granularity of image generation complicates authenticity verification.
Although DMs excel at capturing and recreating images at a semantic level, they are likely less adept at precise, pixel-level reproduction required for intricate visuals such as cartographic materials or detailed technical schematics, due to their inherently stochastic nature.
Notice that the methods discussed in conditional tasks can currently only be achieved at the semantic level, not at the pixel level, leaving pixel-level issues inadequately managed.
In fact, there is a lack of literature focused on the authenticity of content generated by DMs, highlighting a critical need for further research in this area.

\subsubsection{LLMs}

\revise{
LLMs often produce untruthful content, referred to as the hallucination issue, which can be classified into two types: in-context and extrinsic~\citep{respons_14,llm_level3_20}.
In-context hallucination occurs when the output misaligns with the given context, such as providing irrelevant responses or contradicting surrounding information.
Extrinsic hallucination arises when the output conflicts with world knowledge.
The root causes are multifaceted, including reliance on noisy web-crawled datasets \citep{llm_level3_1,llm_level3_2,llm_level3_16}, outdated information, the inherent randomness in top-k decoding methods~\citep{llm_level3_8,llm_level3_9}, and limitations in model and optimization~\citep{llm_level3_3}.
}

\revise{
Early works~\citep{llm_level3_9,second_revision_4,second_revision_12} primarily focused on validating the consistency of LLM outputs with world knowledge.
At a high level, all validation methods first break down the model's responses into atomic facts and then compare them against verified truths to derive a score.
This process is mainly designed for detecting extrinsic hallucination, while the detection of in-context hallucination relies more on the self-correction mechanisms and advanced decoding strategies discussed below.
Next, we review authenticity-enhancing strategies for both the training and inference stages.
}

\revise{
During training, one key strategy is purifying the training datasets.
\citet{llm_level3_10, llm_level3_1} and \citet{llm_level3_11} developed procedures for curating high-quality corpora, while \citet{llm_level3_12} stressed the effectiveness of direct human revisions.
Due to the impracticality of manually revising vast datasets, these studies \citep{llm_level3_15,llm_level3_17,llm_level3_18} proved the potential of a small, high-quality dataset during fine-tuning.
Filtering low-quality texts against reliable language corpora \citep{gpt3} or using LLMs to sift out unreliable information \citep{llm_level3_16} can further enhance the dataset’s integrity.
Additionally, LLMs can be leveraged to generate high-quality data~\citep{llm_level3_19,llm_level3_37,llm_level3_38,llm_level3_39}, and adversarial examples~\citep{llm_level3_40} can serve as high-quality texts.
Another strategy is refining the training process, by either allocating more optimization attention to fact-containing segments~\citep{llm_level3_9} or integrating fact-aware rewards to promote the generation of more reliable content~\citep{second_revision_10}.
}

\revise{
During inference, authenticity can be enhanced through the integration of external knowledge, self-correction mechanisms, and advanced decoding strategies.
RAG allows for accessing reliable, up-to-date information from trusted sources, which can be fused with user queries for contextual clarity \citep{llm_level3_23} or fed into auxiliary models to refine the LLM's original responses \citep{llm_level3_24, llm_level3_25, second_revision_8}.
Self-correction mechanisms capitalize LLMs themselves, with methods forcing them to recall relevant facts \citep{second_revision_7} or prioritizing outputs from attention heads crucial for truthfulness \citep{second_revision_9}.
SelfIE asks LLMs to clarify reasoning and enables fine-grained control over the process \citep{second_revision_11}.
Some approaches \citep{second_revision_5,second_revision_6,llm_level3_28} employ a majority-voting paradigm, which either requires consensus among multiple models or integrates outputs from rephrased queries.
This paradigm also facilitates uncertainty quantification in model responses \citep{llm_level3_27,llm_level3_30}, prompting models to avoid responding when uncertain.
Improved decoding strategies refine reasoning processes or calibrate next-token distribution.
A prime example of the former is "chain of thought," which breaks down complex problems into manageable sub-questions \citep{llm_level3_35}.
This has since evolved into tree-of-thoughts, chain-of-verification, and process supervision \citep{llm_level3_36, llm_level3_33, llm_level3_31}.
For the latter, factual-nucleus sampling \citep{llm_level3_32} adjusts randomness in the top-k sampling process, while context-aware decoding \citep{llm_level3_34} helps LLMs better integrate contextual information and lessen reliance on static, pre-trained knowledge bases.
}

\textbf{Remark.}
Despite these advancements, achieving absolute truthfulness in AI-generated content remains challenging.
Beyond model development, access and information dissemination are also important in practice.
AI service providers should implement stricter usage restrictions and work with platforms to identify AI-generated content, such as requiring “proof of personhood” for users.
Platforms could adopt selective review processes based on the potential impact of the content.
For high-impact content, a hard review may be necessary, requiring users to provide factual sources or proof of identity, along with enlisting administrators or volunteers for manual verification.
For less critical content, a soft review can be employed, which labels the content as "unverified" and conducts reviews only when users raise concerns.
These can be applied to both DMs and LLMs.

\subsection{Benchmark Evaluation Tools: Datasets and Metric}
\label{sec_responsibility_benchmark_tool}
As shown in Table \ref{tab_dataset_metric},
at Level I, the focal point lies in the capacity to differentiate between AI-generated content and human-authored content.
Thus, the principal evaluation metrics encompass the detection rate and AUC.
Advancing to Level II involves the specific challenge of identifying content generated by a specific model.
While the core evaluation metrics remain largely unchanged from Level I, the introduction of watermarks needs additional metrics to assess their impact on content quality.
It should be noted that some watermarking techniques hide specific information within images and Bit Acc serves as a metric to ascertain whether the recovered information matches the written information.

Transitioning to Level III, the evaluation of the authenticity of AI-generated content becomes increasingly intricate.
Diverging from the constrained outputs of discriminative models, generative models produce a spectrum of open-ended results.
Therefore, the assessment of authenticity at this level often requires manual scrutiny, as automated metrics may not sufficiently capture the nuanced variations present in the outputs of generative models.

\subsection{Discussion, Recommendation, and Outlook}

\subsubsection{Discussion}

We review the responsibility of DMs and LLMs from three levels: identifiability, traceability, and verifiability.
For identifiability, DM-generated content is generally easier to detect than LLM-generated content.
Image data captures intricate real-world details, making it difficult for current DMs to achieve a perfect fit.
As a result, AI-generated images exhibit noticeable artifacts resulting from imperfections in the generation process.
In contrast, language data, being a human creation, is structured around rules and patterns, making it easier to fit.
These findings are based on the current landscape, with the caveat that image generation methods are rapidly evolving, and future advancements may overthrow these conclusions.

For traceability, we investigate watermarking techniques.
As the capabilities of generative models reach new heights, differentiating AI-generated content from human-created content becomes increasingly difficult.
This necessitates active intervention in the generation process, such as embedding watermarks for content auditing and promoting responsible usage.
While existing watermarking techniques have proven effective, their resilience against malicious attacks remains questionable.
Attacks on watermarks often involve altering the watermarked content to remove the watermark.
The debate over what constitutes a transformative alteration of AI-generated content into human products is a critical and unresolved issue, deserving further scholarly attention.

Turning to verifiability, the verification of DM-generated content is still in its nascent stages, leaving significant room for exploration.
The definition of authenticity for image data varies across different domains and necessitates further consensus.
Although LLMs have made significant strides in addressing authenticity concerns and proposing potential solutions, their effectiveness is yet to meet expectations and heavily depends on empirical validation.
Consequently, the authenticity of LLM-generated content is currently defensible over tested instances.

\subsubsection{Recommendation}

Based on our review and analysis, we propose the following practical mitigations for practitioners and industry:
\begin{itemize}
    \item Prioritizing responsible strategies, such as watermarking, during foundational design stages is essential for developing responsible AI systems.
    \item Given the current challenges in verifying the authenticity of AI-generated content, a pragmatic approach would involve enabling models to retrieve relevant answers, striking a balance. Alternatively, indicating confidence levels in responses can significantly enhance user caution.
\end{itemize}

\subsubsection{Outlook}

In light of the challenges mentioned above, we suggest exploring the following promising research directions:
\begin{itemize}
    \item Watermark attacks require modifications to AI-generated content. However, determining the extent of alteration that renders content unrecognizable as AI-generated remains an open question. Further studies are needed to establish clear boundaries for when rewritten material loses its original AI authorship.
    \item The authenticity of images generated by DMs is largely unexamined. This issue is critical in fields requiring high precision but has not received sufficient attention from academia or industry, possibly due to the current focus on using DMs for artistic tasks.
    \item While substantial research has been conducted to enhance the authenticity of LLM-generated content, empirical validation alone is inadequate. Instead, it is vital to identify worst-case scenarios or establish theoretical frameworks for assessing authenticity.
\end{itemize}

\section{Conclusion}
\label{section:conclusion}

We reviewed recent developments regarding the trustworthiness of these models through the lenses of privacy, security, fairness, and responsibility.
To promote the trustworthy usage of these models and mitigate associated risks, we proposed some actionable steps for both companies and users.
Additionally, we identified challenges that the research community should address.
Through these efforts, we seek to improve the understanding and management of the risks associated with DMs and LLMs, ensuring their reliable deployment for the benefit of society as a whole.

\section*{Acknowledgements}
This work was supported by the National Natural Science Foundation of China under grant number 62202170, Alibaba Group through Alibaba Innovative Research Program, and the Open Research Fund of KLATASDS-MOE, ECNU.

\section*{Data Availability}
This paper serves as a survey and does not include datasets.

\bibliographystyle{spbasic}
{\footnotesize
\bibliography{refs}}

\end{document}